%% file: root.tex
\RequirePackage[T1]{fontenc}
\documentclass[letterpaper,10pt,conference]{ieeeconf}
\IEEEoverridecommandlockouts
\usepackage{amsmath,amssymb,graphicx,booktabs,tabularx,multirow,float,array}
\let\labelindent\relax
\usepackage{enumitem}
\usepackage{etoolbox}
\usepackage{cite}
\let\citep\cite
\usepackage{url}
\usepackage[hidelinks]{hyperref}
\usepackage[table]{xcolor}
\newcolumntype{C}[1]{>{\centering\arraybackslash}p{#1}}
\newif\ifDAVISappendix
\DAVISappendixfalse
\makeatletter
\let\DAVISoriginalcaption\@makecaption
\long\def\@makecaption#1#2{%
  \ifx\@captype\@IEEEtablestring
    \par\vspace{3pt}%
    {\footnotesize\normalfont
      \sbox\@tempboxa{#1: #2}%
      \ifdim\wd\@tempboxa<\linewidth
        \centering #1: #2\par
      \else
        \noindent #1: #2\par
      \fi}%
    \ifDAVISappendix\vspace{4pt}\fi%
  \else
    \DAVISoriginalcaption{#1}{#2}%
  \fi}
\makeatother
\usepackage{capt-of}
\usepackage{cuted}

\graphicspath{{figures/}{figures/teaser/}{figures/system/}{figures/method/}{figures/experiments/}{figures/real_world/}}

\newcommand{\AuJK}[1]{#1}

\newcommand{\ZXX}[1]{#1}
\definecolor{addcol}{RGB}{150,20,140}

\title{\LARGE\bfseries DAVIS: A \textbf{\textcolor{red}{D}}epth-Only End-to-End \textbf{\textcolor{red}{A}}ctive-\textbf{\textcolor{red}{V}}ision Framework for Humano\textbf{\textcolor{red}{I}}d \textbf{\textcolor{red}{S}}occer Skills}
\author{%
  \textbf{Jiakang Jin$^{1,*}$, Yixiao Huo$^{1,2,*}$, Pengyuan Wang$^{1,*}$, Yinan Han$^{1,*}$,}\\
  \textbf{Tingxuan Zhang$^{1}$, Zhuobing Zhao$^{1}$, Xuanxin Zhou$^{1}$, Zhangchen Ye$^{1,2}$, Enxuan Ruan$^{1}$,}\\
  \textbf{Yifei Bao$^{1}$, Jiankun Yang$^{1}$, Chenghao Sun$^{1}$, Wenhao Cui$^{1}$, Xiaoyu Tian$^{1,\dagger}$, Yiming Li$^{2,\dagger}$}\\[2pt]
  \small $^{1}$Noetix Robotics \quad $^{2}$Tsinghua University\\
  \small $^{*}$Equal Contribution \quad $^{\dagger}$Corresponding Author\\[2pt]
  \small Project Page: \url{https://thusi-lab.github.io/DAVIS/}%
}

\begin{document}
\maketitle
\thispagestyle{empty}
\pagestyle{empty}

\begin{strip}
  \centering
  \begin{minipage}{\textwidth}
    \centering
    \vspace{-3em}
    \includegraphics[width=\textwidth]{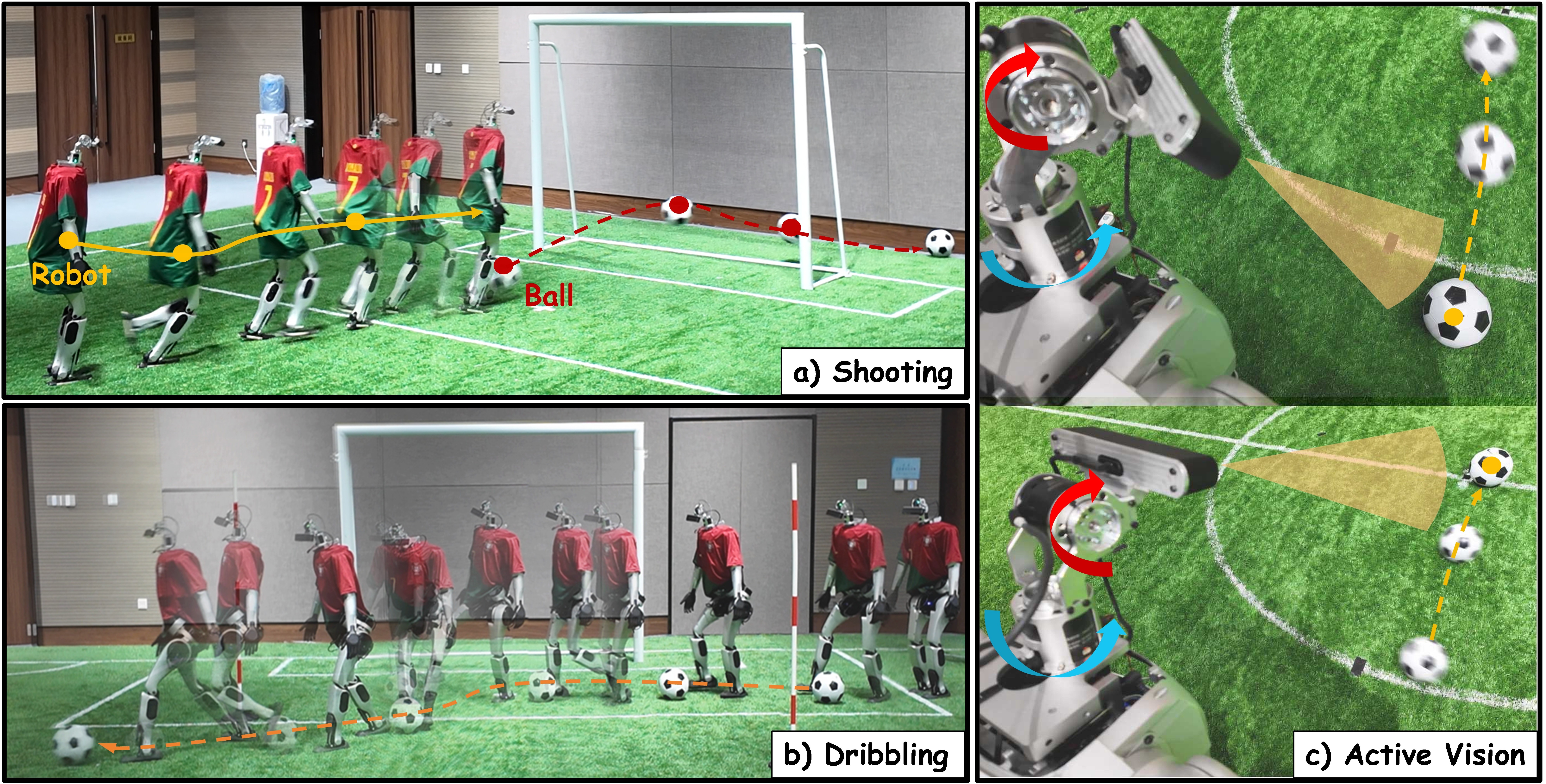}
    \vspace{-2em}
    \captionof{figure}{\textbf{Overview of our humanoid soccer system.}
    (a) The robot performs goal-oriented shooting by approaching the ball and kicking it into the goal.
    (b) The robot demonstrates agile multi-directional dribbling with continuous ball control, including turning and weaving around poles.
    (c) Our active vision design enables the robot to actively reorient its head in yaw and pitch to keep the ball near the center of the field of view, thereby improving perception-action coordination.}
    \vspace{-1em}
    \label{fig:teaser}
  \end{minipage}
\end{strip}

\begin{abstract}
\input{sections/abstract.tex}

\end{abstract}

\section{Introduction}\label{sec:introduction}

\input{sections/intro.tex}

\section{Related Work}\label{sec:related-work}

\input{sections/related_work.tex}

\begin{figure*}[!t]
  \centering
  \includegraphics[width=\linewidth]{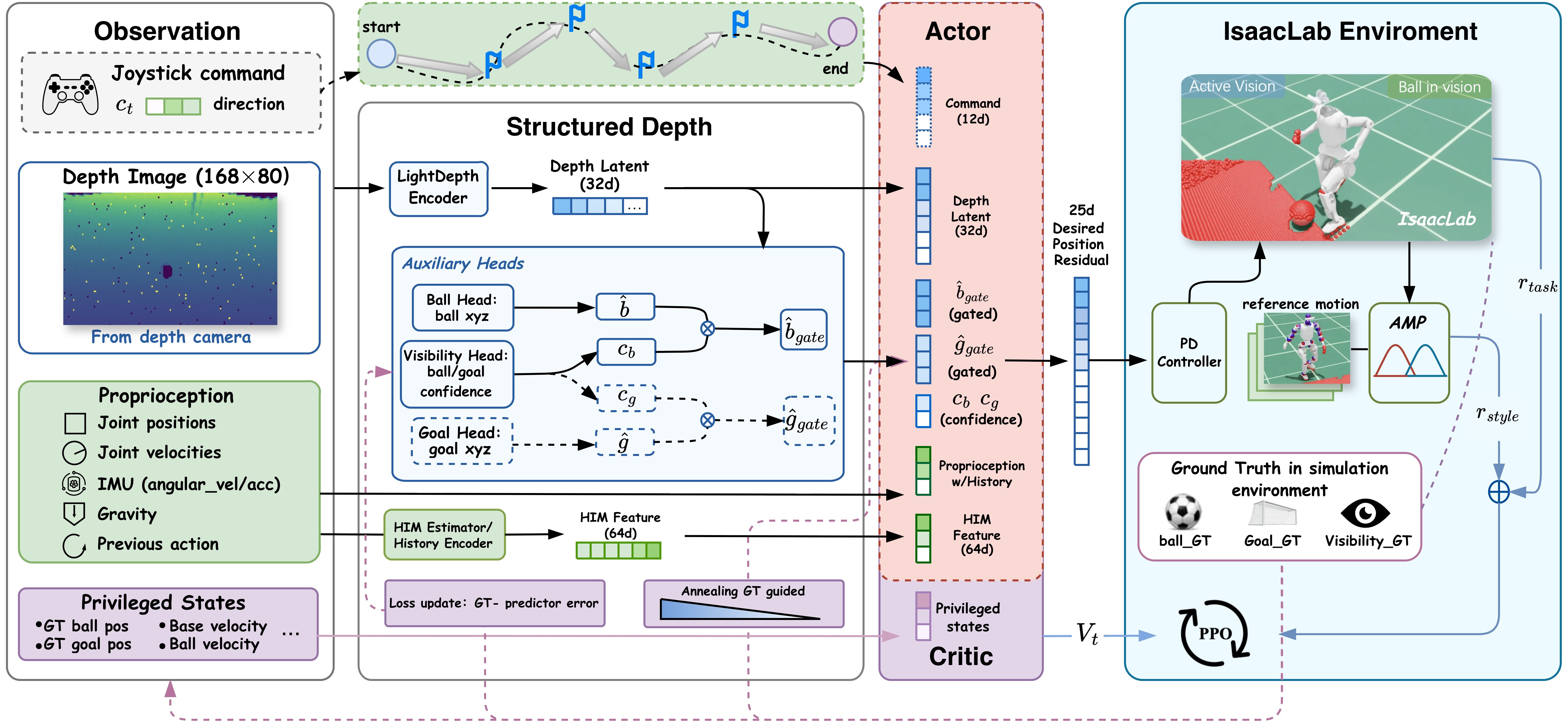}
  \vspace{-2em}
  \caption{\textbf{Overview of our DAVIS framework.} The observation stream contains the command $c_t$, \textbf{a single $168 \times 80$ depth image}, proprioceptive history, and privileged simulation states. The structured-depth module encodes the depth input into a 32-D latent, predicts ball/goal geometry and visibility confidence through \textbf{auxiliary heads}, forms confidence-gated geometry features, and combines them with a 64-D HIM/history feature. The actor receives the command, depth latent, gated geometry/confidence features, proprioceptive history, and HIM feature, and outputs a 25-D position residual for the PD controller. The auxiliary losses, GT-to-prediction annealing, PPO update, task reward, and style reward use simulation-only signals from IsaacLab rollouts, including ball, goal, and visibility ground truth. Dashed boxes in the structured-depth module indicate optional task-specific modules.}
  \vspace{-1em}
  \label{fig:framework}
  
\end{figure*}

\section{Method}\label{sec:method}

\subsection{Problem Formulation}\label{sec:problem-formulation}

\input{sections/method/problem_formulation.tex}

\subsection{Depth-Only End-to-End Structured Soccer RL Framework}\label{sec:framework}

\input{sections/method/framework.tex}

\subsection{Visibility-Aware Auxiliary Geometry}\label{sec:visibility-aux}

\input{sections/method/visibility.tex}

\begin{figure*}[!t]
  \centering
  \includegraphics[width=\textwidth]{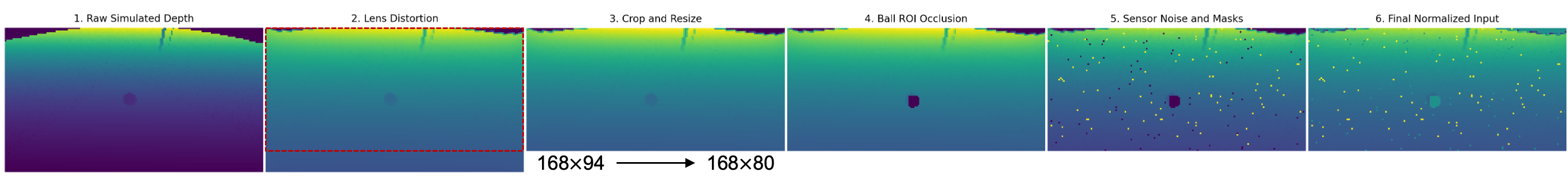}
  \vspace{-2em}
  \caption{Depth alignment transforms the raw depth frame into the aligned policy input through distortion correction, cropping from $168\times94$ to $168\times80$, ball ROI occlusion, sensor noise, random masks and normalization, so the same single-frame depth interface can be used both in sim and real.}
  \vspace{-0.8em}
  \label{fig:pipeline}
  
\end{figure*}

\subsection{Skill Instantiation}\label{sec:skill-instantiations}

\input{sections/method/skill_instantiations.tex}

\subsubsection{Goal-Directed Shooting}\label{sec:shooting-instantiation}

\input{sections/method/shooting.tex}

\subsubsection{Directional Dribbling}\label{sec:dribbling-instantiation}

\input{sections/method/dribbling.tex}

\begin{figure*}[!t]
  \centering
  \includegraphics[width=\linewidth]{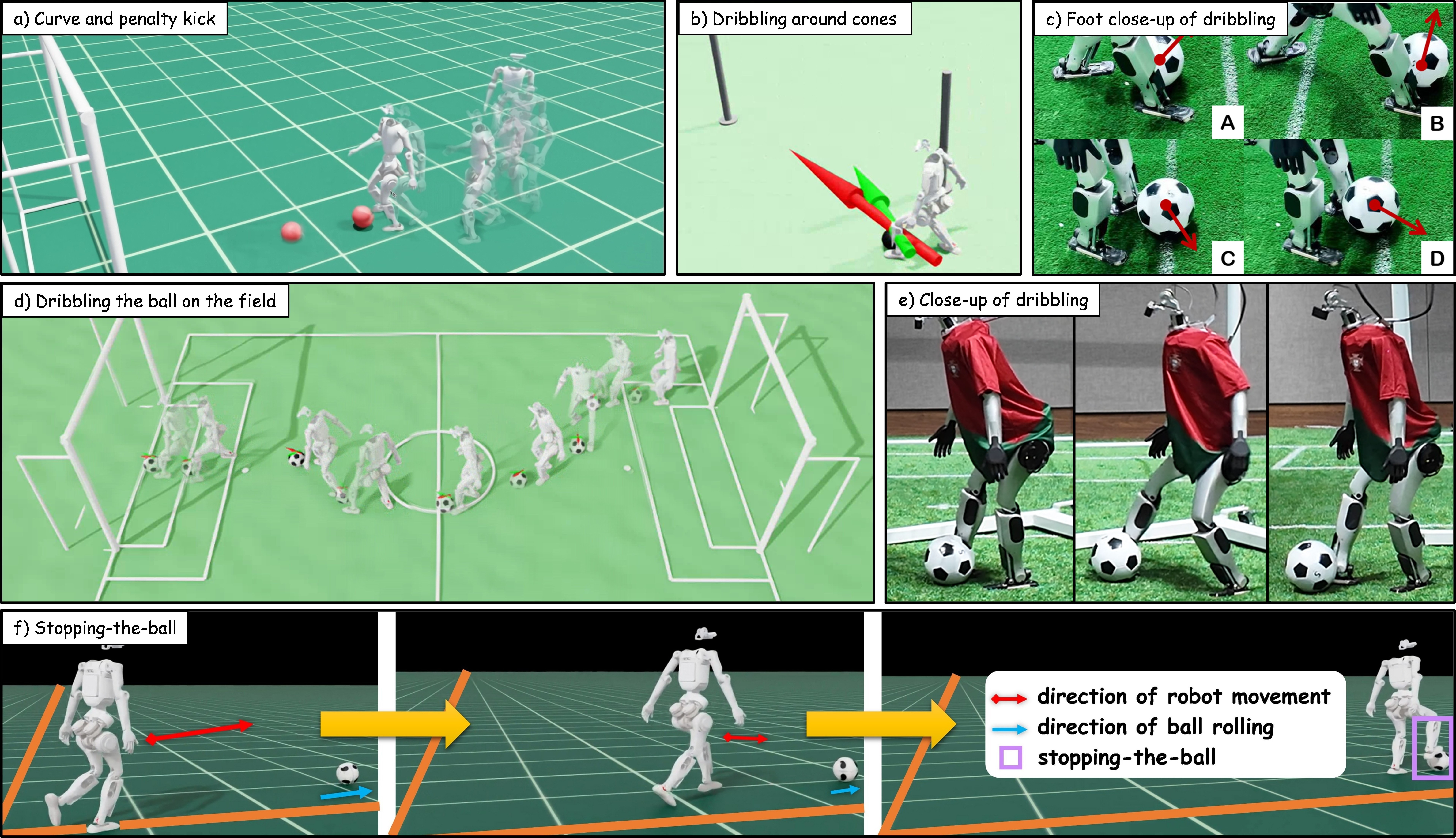}
  \vspace{-2em}
  \caption{\AuJK{\textbf{Qualitative examples of learned soccer skills.} In simulation: (a) curved and penalty kicks, (b) cone-avoidance dribbling, (d) dribbling across a soccer field, and (f) stopping a rolling ball, where red and blue arrows indicate robot and ball motion, respectively, and the purple box marks the stopped ball. On the real robot: (c) foot--ball contact during dribbling, with arrows indicating ball motion, and (e) a close-up dribbling sequence.}}
  \vspace{-0.8em}
  \label{fig:qualitative-soccer}
  
\end{figure*}


\subsection{Training and Deployment Pipeline}\label{sec:training-deployment}

\input{sections/method/pipeline.tex}

\begin{table*}[!t]
  \centering
  \scriptsize
  \setlength{\tabcolsep}{1.5pt}
  \renewcommand{\arraystretch}{0.90}

  \caption{Quantitative results for goal-directed shooting and directional dribbling
  in simulation and on the real robot. Shooting entries and real-robot dribbling
  entries report success rate (SR) on a 0--1 scale; simulation dribbling reports SR,
  velocity error ($E_{\mathrm{vel}}$), and ball-in-band ratio ($R_{\mathrm{bib}}$)
  over 900 repeated-S slalom trials. Arrows indicate the preferred direction, and
  bold marks the best result within each fixed-$N$ simulation-dribbling group. For shooting, a success requires the ball to cross the goal line in one kick. For dribbling, a success requires the robot to remain upright throughout execution and both the robot and ball to finish within 1\,m of the target.}
  \vspace{0.5em}
  \label{tab:quantitative-results}
  \begin{tabularx}{\textwidth}{@{}C{0.9cm} C{1.8cm} C{1.35cm} C{1.35cm} C{1.35cm} C{0.65cm}
                                 C{0.75cm} C{1.15cm}
                                 >{\centering\arraybackslash}X
                                 >{\centering\arraybackslash}X
                                 >{\centering\arraybackslash}X
                                 >{\centering\arraybackslash}X
                                 >{\centering\arraybackslash}X
                                 >{\centering\arraybackslash}X
                                 >{\centering\arraybackslash}X
                                 >{\centering\arraybackslash}X
                                 >{\centering\arraybackslash}X@{}}
    \toprule
    \multicolumn{6}{c}{\textbf{Goal-Directed Shooting}}
      & \multicolumn{11}{c}{\textbf{Directional Dribbling}} \\
    \cmidrule(lr){1-6}\cmidrule(lr){7-17}

    \multirow{2}{*}{\centering Setting} & \multirow{2}{*}{\centering Metric}
      & \multirow{2}{*}{\centering Easy} & \multirow{2}{*}{\centering Med.}
      & \multirow{2}{*}{\centering Hard} & \multirow{2}{*}{\centering Total}
      & \multirow{2}{*}{\centering Setting} & \multirow{2}{*}{\centering Metric}
      & \multicolumn{3}{c}{$N=3$}
      & \multicolumn{3}{c}{$N=5$}
      & \multicolumn{3}{c}{$N=7$} \\
    \cmidrule(lr){9-11}\cmidrule(lr){12-14}\cmidrule(lr){15-17}

    & & & & & & & &
      $60^\circ$ & $90^\circ$ & $120^\circ$
      & $60^\circ$ & $90^\circ$ & $120^\circ$
      & $60^\circ$ & $90^\circ$ & $120^\circ$ \\
    \midrule

    \multirow{3}{*}{\centering Sim.}
      & Penalty SR $\uparrow$
      & 0.95 & 0.96 & 0.56 & 0.85
      & \multirow{3}{*}{\centering Sim.}
        & SR $\uparrow$
        & 0.71 & \textbf{0.77} & 0.65
        & 0.66 & \textbf{0.72} & 0.55
        & \textbf{0.68} & 0.64 & 0.43 \\

      & Free-kick SR $\uparrow$
      & 1.00 & 0.88 & 0.71 & 0.85
      & & $E_{\mathrm{vel}}\downarrow$
        & \textbf{0.69} & 0.74 & 0.91
        & 0.74 & \textbf{0.73} & 0.80
        & \textbf{0.71} & 0.81 & 0.97 \\

      & --
      & -- & -- & -- & --
      & & $R_{\mathrm{bib}}\uparrow$
        & 0.73 & \textbf{0.74} & 0.64
        & \textbf{0.76} & 0.75 & 0.70
        & \textbf{0.76} & 0.72 & 0.60 \\

    \cmidrule(lr){1-6}\cmidrule(lr){7-17}

    Real
      & Penalty SR $\uparrow$
      & 0.68(26/38) & 0.57(28/49) & 0.55(33/60) & {--}
      & Real & SR $\uparrow$
        & \multicolumn{3}{c}{Easy: 0.79(22/28)}
        & \multicolumn{3}{c}{Medium: 0.68(19/28)}
        & \multicolumn{3}{c}{Hard: 0.61(17/28)} \\
    \bottomrule
  \end{tabularx}
  \vspace{-1em}

\end{table*}

\section{Experiments}\label{sec:experiments}

\input{sections/experiments/overview.tex}

\begin{figure*}[!t]
    \centering
    \includegraphics[width=\textwidth]{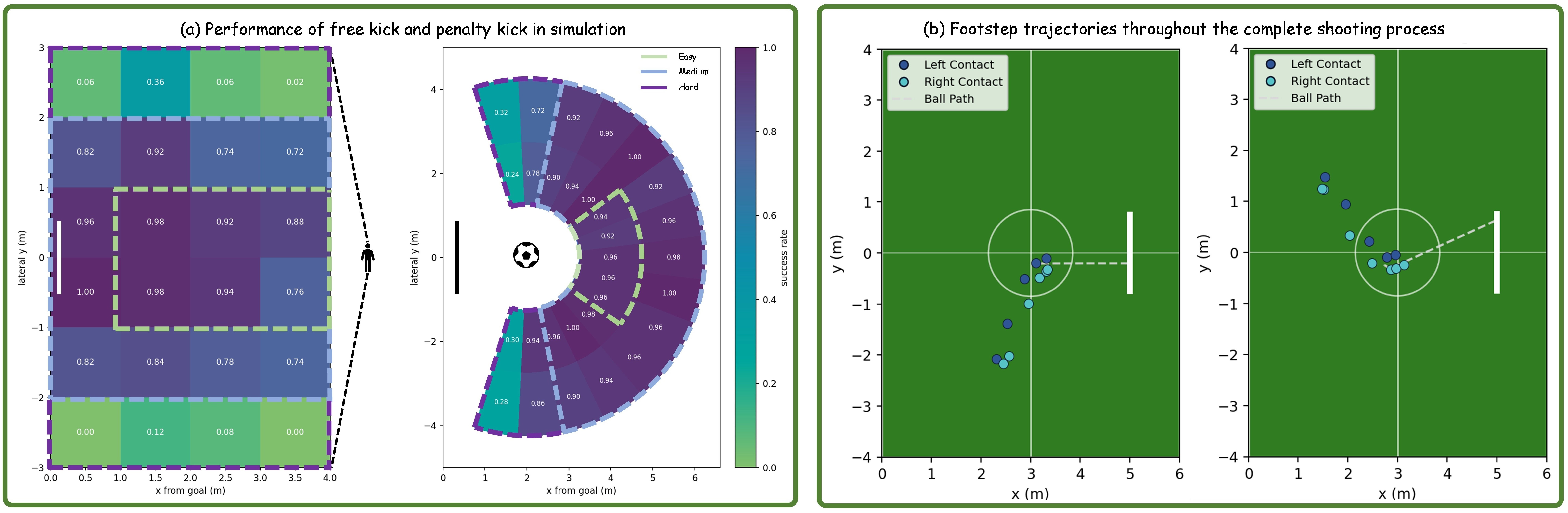}
    \vspace{-2em}
    \caption{ (a) Free-kick and Penalty-kick performance in simulation. Each region uses over 50 trials. Darker colors indicate higher success rates, and dashed lines mark difficulty regions. (b) Footstep trajectories during shooting. The policy shows a smooth gait and consistent speed pattern. }
    \vspace{-0.8em}
    \label{fig:shooting_results}
    
\end{figure*}

\begin{figure*}[!t]
  \centering
  \includegraphics[width=\textwidth]{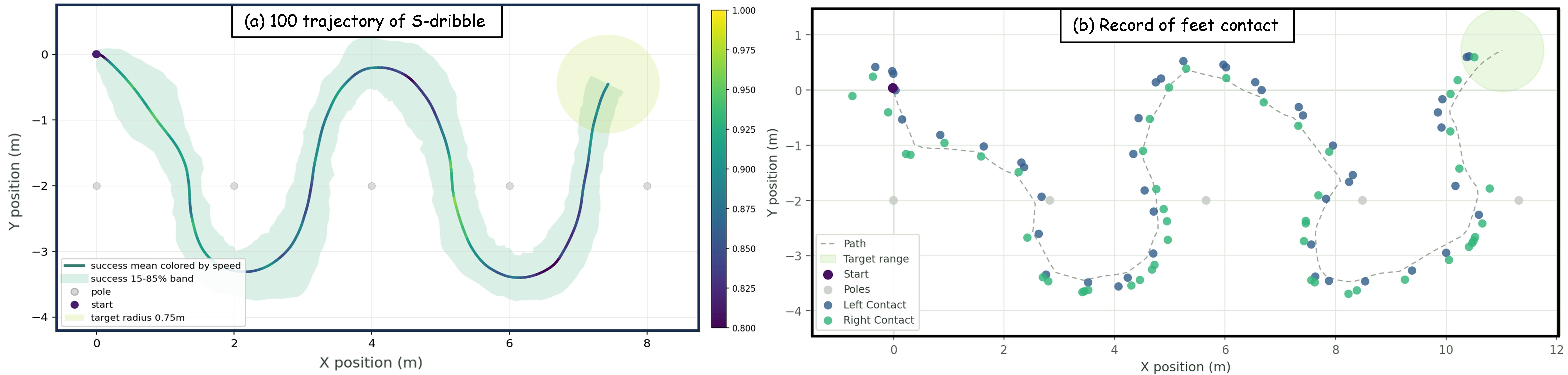}
  \vspace{-2em}
  \caption{Performance of repeated S-slalom in simulation. (a) Statistical results of 100 S-slalom trajectories. (b) Foot-contact record from a representative successful trajectory.}
  \vspace{-2em}
  \label{fig:dribble-sim}
  
\end{figure*}

\subsection{Simulation Experiments}\label{sec:simulation}

\input{sections/experiments/simulation.tex}

\subsection{Ablation Study}\label{sec:ablation}

\input{tables/ablation/unified_ablation.tex}

\input{sections/experiments/ablation.tex}

\subsection{Real-World Deployment}\label{sec:real-world}

\input{sections/experiments/real_world.tex}

\section{Conclusion}\label{sec:conclusion}

\input{sections/conclusion.tex}

\section{Limitations and Future Work}\label{sec:limitations}

\input{sections/limitations.tex}

\bibliographystyle{IEEEtran}
\bibliography{bibliography/references}

\DAVISappendixtrue
\AtBeginEnvironment{table}{\scriptsize}
\AtBeginEnvironment{table*}{\scriptsize}
\renewcommand{\tabularxcolumn}[1]{m{#1}}
\renewcommand{\arraystretch}{0.96}
\appendix
\section{Supplementary Material}
\label{sec:appendix}
\input{appendix/appendix.tex}

\end{document}

%% file: sections/abstract.tex
Humanoid soccer contact skills require more than producing high-impact foot--ball contacts: the robot must close the loop over perception, approach, alignment, impact, and recovery while its own motion induces substantial viewpoint changes, frequent loss of the ball from view, and uncertain contact outcomes. In this work, we ask a compact yet stricter question: can a humanoid learn soccer contact skills using only a head-mounted depth image, proprioceptive history, and an optional low-dimensional task command, and directly output 25-DoF joint PD targets without extra runtime perception or planning modules? To this end, we propose \textbf{DAVIS}, a depth-only end-to-end framework for humanoid soccer skills that learns visibility-aware auxiliary geometry during training, and combines GT-to-prediction annealing, task curricula, and AMP-style motion priors to smoothly bridge privileged supervision and real deployment. Built on this framework, we instantiate representative soccer contact skills, including goal-directed shooting and directional dribbling, through task-specific definitions of objects, commands, rewards, and curricula, and validate them through simulation, Noetix E1 real-robot experiments, and ablations.


%% file: sections/intro.tex
Humanoid robots have made substantial progress in reinforcement-learning-driven dynamic locomotion~\cite{rudin2022learning,miki2022perceptive,zhuang2024humanoidparkour,he2025asap,zhu2026hiking,wu2026php}, but soccer remains a harder setting because it requires the robot to perceive, approach, align, contact, and recover while the \AuJK{target and the robot itself are moving}. Unlike standard locomotion benchmarks, soccer contact skills require the policy to reason not only about \textit{where the ball is}, but also about \textit{how to approach it}, \textit{how to keep the target in view}, and \textit{how contact should change the subsequent ball motion}.

On real humanoids, onboard sensing and whole-body control further amplify these difficulties: the head-mounted view changes with body motion, active head yaw and pitch determine what evidence the policy receives, and small perception or stance errors can change the contact outcome~\cite{xiong2025via,liu2025avr,cheng2024opentelevision,nakagawa2025active_neck,fu2025demohlm}. Existing soccer systems often manage this complexity with modular perception, localization, planning, and skill-switching interfaces, including RGB detectors, LiDAR/RGB-D-based localization, or external state inputs. \AuJK{Because soccer contact skills depend on local ball and goal geometry, head-mounted depth provides a direct geometric observation from which to study control without externally supplied scene state. This motivates the question:} \textbf{{can a humanoid learn deployable soccer contact skills from only head-mounted depth, proprioceptive history, and low-dimensional task commands --- without any external scene state?}}

\AuJK{We propose \textbf{DAVIS}, a depth-only end-to-end reinforcement learning framework to answer this question.} At deployment, the policy remains \AuJK{a closed-loop mapping} from head-mounted depth, proprioceptive history, and low-dimensional commands to 25-DoF joint PD targets. This is achieved by placing structure inside the learning signal rather than in the deployment pipeline, following the \AuJK{idea} of using privileged simulation information to guide deployable policies~\cite{kumar2021rma,wang2024cts,zhang2025slim}. \AuJK{Auxiliary geometry and visibility predictions are learned from privileged simulation supervision, while the actor gradually transitions from stable geometric guidance to self-predicted features via GT-to-prediction annealing and visibility gating.} With this interface fixed, different soccer behaviors can be learned \AuJK{by changing} the local objective, geometry, and supervision that define the task.

DAVIS is designed to support a range of soccer skills through a shared learning framework: new skills are specified by task objects, commands, rewards, and curricula while retaining the depth-to-control interface. We evaluate goal-directed shooting and directional dribbling in detail as representative instances of one-shot striking and sustained ball control, using them to assess the feasibility of the framework across complementary contact regimes. Additional examples, including stopping the ball and obstacle-aware dribbling, are shown in Fig.~\ref{fig:qualitative-soccer}; ball-loss recovery is demonstrated in the project-page video rather than as a panel in Fig.~\ref{fig:qualitative-soccer}. Together, these instances provide evidence of the framework's extensibility through task-specific training.

In summary, our contributions are as follows: (1) We introduce a \textbf{depth-only end-to-end RL framework} for humanoid soccer contact skills, where the deployed policy maps head-mounted depth, proprioceptive history, and low-dimensional task commands directly to 25-DoF joint targets with active-vision head control. (2) We develop a \textbf{visibility-aware auxiliary geometry mechanism} that learns task-relevant object geometry from the depth latent, and combines visibility masking with GT-to-prediction annealing to stabilize visual learning without adding runtime perception modules. (3) \AuJK{We instantiate multiple soccer skills within the same framework and establish a complete pipeline from IsaacLab training to E1 real-robot deployment, including depth alignment, policy export, and 25-DoF PD control. Real-robot evaluations and additional skill demonstrations support the framework's practical feasibility and applicability across soccer tasks.}

%% file: sections/related_work.tex
\subsection{\AuJK{Robot Soccer Skill Learning}}

\input{sections/related_work/humanoid_soccer.tex}

\subsection{\AuJK{Whole-Body Loco-Manipulation}}
\input{sections/related_work/coupled_whole-body_control_for_dynamic_loco-manipulation.tex}

%% file: sections/related_work/humanoid_soccer.tex
Robot soccer has long served as a challenging benchmark for robot control, requiring the integration of visual perception, real-time decision making, agile locomotion, and ball-oriented skills~\cite{kitano1997robocup,gerndt2015humanoid,beukman2024robocupgym}. 
RoboCup-style soccer systems often relied on modular perception-control pipelines, where sensing, state estimation, and soccer control were separately engineered rather than learned end-to-end~\cite{yi2016hierarchical}.
Recent learning-based methods have enabled legged robots to acquire soccer skills through various paradigms, including reinforcement learning and imitation learning
~\cite{leottau2015ball,dasilva2021deep_rl_soccer,abreu2025skilled_soccer_team,ji2022soccer_shooting,huang2023goalkeeper,ji2023dribblebot,haarnoja2024agile_soccer,tirumala2025egocentric_soccer}. 
Yet most existing systems still depend on RGB-based object detection or multi-sensor perception pipelines~\cite{wang2025dribblemaster,kong2026paid,ye2026skillx,wang2025visiondriven}.

In contrast, our work studies depth-driven humanoid soccer skill learning, where the robot learns to chase, dribble, and kick \AuJK{using only depth observations and proprioception}.

%% file: sections/related_work/coupled_whole-body_control_for_dynamic_loco-manipulation.tex
Whole-body loco-manipulation on humanoid robots is inherently challenging, as the robot must simultaneously coordinate balance, locomotion, and task-directed manipulation within a high-dimensional and tightly coupled control space~\citep{sentis2006wholebody,herzog2016momentum,he2024hover}. 
To make the problem tractable, prior systems often introduce decoupled or hierarchical interfaces, such as separate upper- and lower-body controllers r high-level task policies paired with low-level whole-body controllers~\citep{li2025softa,ben2025homie,xue2025hugwbc,sun2025ulc}. 
Humanoid soccer further amplifies this difficulty: unlike manipulating static objects, the ball is a highly dynamic target, requiring the robot to approach from a suitable direction, maintain visibility, and generate precise foot--ball contacts for effective ball control. 

\AuJK{To address these challenges,} we propose a whole-body control framework, \textbf{DAVIS}, that jointly optimizes active visual tracking, approach, and foot--ball interaction for dynamic ball-centric loco-manipulation.

%% file: sections/method/problem_formulation.tex
We model humanoid soccer contact skills as a POMDP problem ~\cite{kaelbling1998planning}, denoting the process as $\mathcal{M}=\langle\mathcal{S},\mathcal{A},\mathcal{T},\mathcal{R},\gamma\rangle$. The simulator maintains a full state $\mathbf{s}_t\in\mathcal{S}$ containing robot, ball, goal, and contact-related physical variables; during training, privileged variables derived from this state, such as ball pose, goal pose, visibility, contact state, and reference motion, are used as critic inputs, reward-computation terms, and auxiliary-supervision labels. The deployed policy instead observes only a head-mounted depth image $\mathbf{D}_t$, a proprioceptive history $\mathbf{p}_{t-H:t}$ of length $H$, and an optional low-dimensional task variable $\mathbf{c}_t$; for instances without an external command, $\mathbf{c}_t$ is omitted. We write the deployed observation and action as $\mathbf{o}_t=(\mathbf{D}_t,\mathbf{p}_{t-H:t},\mathbf{c}_t)$ and $\mathbf{a}_t=\pi_\theta(\mathbf{o}_t)\in\mathbb{R}^{25}$. The proprioceptive history is encoded inside the policy to infer latent body dynamics, while all explicit task geometry used by the actor must be produced from deployable inputs. The low-level controller converts the action into joint PD targets $\mathbf{q}^{\mathrm{des}}_t=\mathbf{q}^0+\mathbf{s}\odot\mathbf{a}_t$, where $\mathbf{q}^0$ is the nominal joint configuration and $\mathbf{s}$ is the per-joint action scale. The E1 humanoid consists of 23 body DoFs and 2 head DoFs, so we decompose the action as $\mathbf{a}_t=[\mathbf{a}_t^{\mathrm{body}},\mathbf{a}_t^{\mathrm{head}}]$ with $\mathbf{a}_t^{\mathrm{body}}\in\mathbb{R}^{23}$ and $\mathbf{a}_t^{\mathrm{head}}\in\mathbb{R}^{2}$. Because the depth camera is mounted on the actuated head, the head action is part of the sensing loop: it directly changes the next depth observation. We optimize the asymmetric actor-critic policy with PPO~\cite{schulman2017ppo} by maximizing the expected discounted return
\begin{equation}\label{eq:ppo-objective}
\max_\theta\mathbb{E}_{\pi_\theta}\!\left[\sum_{t=0}^{T}\gamma^t\left(r_t^{\mathrm{task}}+\lambda_{\mathrm{amp}}r_t^{\mathrm{amp}}-p_t^{\mathrm{reg}}\right)\right],
\end{equation}
\AuJK{where $r_t^{\mathrm{task}}$ denotes task reward, $r_t^{\mathrm{amp}}$ encourages motion near a reference prior, and $p_t^{\mathrm{reg}}$ penalizes joint limits, action changes, unstable posture, collisions, and unsafe contact.}

%% file: sections/method/framework.tex
We build \textbf{DAVIS} around a deployment interface that remains deliberately narrow: the actor receives head-mounted depth, proprioceptive history, and an optional low-dimensional command, and outputs 25-DoF joint target residuals for the PD controller, as illustrated in Fig.~\ref{fig:framework}. During training, privileged simulation states provide dense supervision for perception, value learning, and motion regularization; at deployment, the policy keeps only the depth/proprioception-to-control path.

To keep this deployment path compact, we introduce task structure only through training-time supervision and regularization. The depth latent feeds auxiliary geometry and visibility heads, whose predictions are confidence-gated before entering the actor. Simulation ground truth supplies geometry and visibility labels, but only as auxiliary supervision and as early-stage geometric guidance. A GT-to-prediction annealing schedule gradually replaces privileged features with predicted features, so the actor first learns how local geometry should affect whole-body motion and then operates with geometry inferred from depth. In parallel, PPO optimizes task return, the history estimator learns proprioceptive state inference, and AMP-style motion priors regularize full-body motion. These components make long-horizon soccer contact learning tractable while preserving a single deployable policy graph~\cite{peng2018deepmimic,peng2021amp,escontrela2022adversarial}.

The next subsection specifies how task objects, variables, rewards, curricula, and optional contact conditioning instantiate individual skills while preserving this deployment interface.

%% file: sections/method/visibility.tex
\begin{figure}[t]
  \centering
  \includegraphics[width=\linewidth]{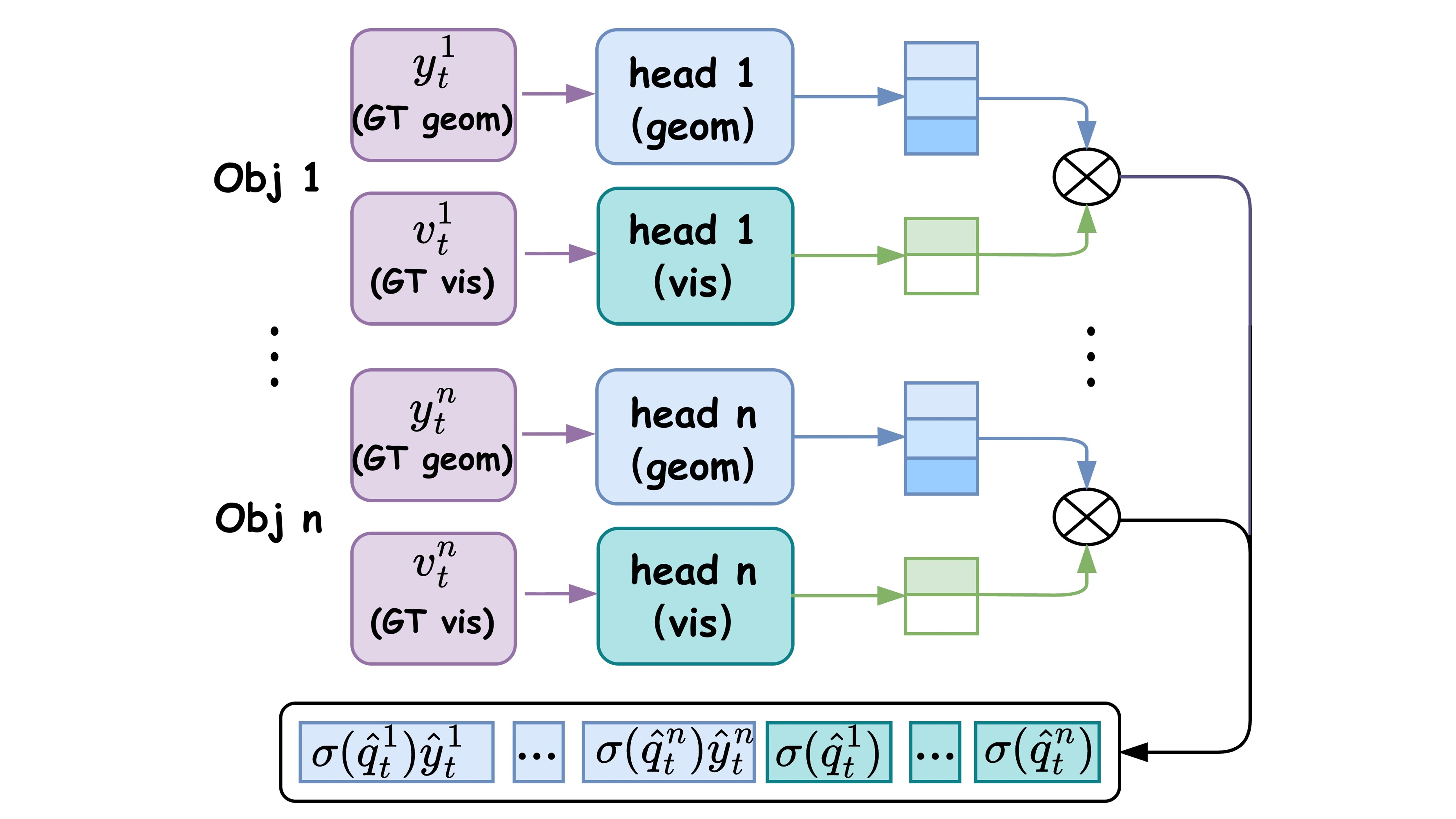}
  \vspace{-2em}
  \caption{Scalable auxiliary geometry. Each task object is assigned a geometry head and a visibility head supervised by privileged labels during training; their outputs are visibility-gated and concatenated into the actor feature, enabling the same mechanism to extend from one object to $n$ objects.}
  \vspace{-1em}
  \label{fig:auxiliary-geometry}
  
\end{figure}

Learning task geometry from depth only through task returns is inefficient: task objects can occupy a small image region, while the gradient from PPO returns must pass through the actor and the depth encoder before shaping visual features. We therefore attach a visibility-aware auxiliary geometry module to the depth latent and supervise it with privileged simulation labels during training. \AuJK{As shown in Fig.~\ref{fig:auxiliary-geometry}, the module predicts object-level geometry and visibility inside the policy, so the actor receives direct object-level supervision without adding a runtime perception module.}

Let $\mathbf{z}_t^D=f_D(\mathbf{D}_t)\in\mathbb{R}^{32}$ be the depth latent.
\AuJK{For each task-relevant object $\mathrm{obj}_i\in\mathcal{O}$, the auxiliary module predicts its 3-D center and visibility confidence from the depth latent,}
\begin{equation}\label{eq:aux-prediction}
(\hat{\mathbf{y}}_t^i,\hat{q}_t^i)
=
h_{\mathrm{aux}}^i(\mathbf{z}_t^D),
\qquad
\hat{p}_t^i=\sigma(\hat{q}_t^i),
\end{equation}
where
$\mathbf{y}_t^i=[x_t^i,y_t^i,z_t^i]^\top$
denotes the object center in a right-handed head-yaw-aligned frame, with
$x$ along the head-yaw forward direction, $y$ to the left, and $z$ upward.
Both geometry and visibility predictions depend only on the current 32-D depth latent; head states are provided separately to the actor through proprioception.
The same construction extends to
$\mathcal{O}=\{\mathrm{obj}_i\}_{i=1}^{n}$
by assigning each object an indexed 3-D center prediction
$\hat{\mathbf{y}}_t^i$, visibility prediction $\hat{q}_t^i$,
ground-truth center $\mathbf{y}_t^i$, and visibility label $v_t^i$.

Visibility is learned jointly with geometry because a head-mounted camera can temporarily lose task-relevant objects.
Geometry is supervised only when the corresponding object is visible, while visibility is trained with a binary objective,
\begin{equation}\label{eq:aux-loss}
\mathcal{L}_{\mathrm{aux}}
=
\sum_{i=1}^{n}
\lambda_i v_t^i
\ell_i(\hat{\mathbf{y}}_t^i,\mathbf{y}_t^i)
+
\sum_{i=1}^{n}
\lambda_{v_i}
\mathrm{BCE}(\hat{q}_t^i,v_t^i).
\end{equation}
Visibility labels are computed using the full camera transform together with valid field-of-view and depth-range constraints; for multi-point objects, visibility is positive if any predefined structural point is valid, while the geometry target always remains the single 3-D object center.

To avoid forcing the actor to rely on unreliable predictions early in training, we gradually replace privileged geometry with predicted geometry.
Let $\beta_k\in[0,1]$ denote the prediction weight at policy update $k$.
For object $i$,
\begin{equation}\label{eq:geometry-anneal}
\mathbf{y}_{t,k}^{e,i}
=
(1-\beta_k)\mathbf{y}_t^i
+
\beta_k\hat{\mathbf{y}}_t^i .
\end{equation}
Using the correspondingly annealed visibility confidence $p_{t,k}^{e,i}$, we define the detached confidence gate
\begin{equation}\label{eq:visibility-gate}
c_{t,k}^i
=
\operatorname{clip}
\left(
\operatorname{stopgrad}(p_{t,k}^{e,i}),
0.05,1
\right).
\end{equation}
The actor-side auxiliary representation is then
\begin{equation}\label{eq:aux-feature}
\mathbf{u}_{t,k}^{\mathrm{aux}}
=
\operatorname{Concat}_{i=1}^{n}
\left[
c_{t,k}^i\mathbf{y}_{t,k}^{e,i},
c_{t,k}^i
\right].
\end{equation}
\AuJK{As $\beta_k$ increases from $0$ to $1$, the actor transitions from privileged geometric guidance to depth-based predictions.}
At deployment, all auxiliary features are predicted solely from depth.

%% file: sections/method/skill_instantiations.tex
Once the deployment interface in Sec.~\ref{sec:framework} is fixed, a soccer skill is specified by a task definition rather than by switching runtime modules. \AuJK{This separation makes the interface reusable: new skills retain the depth-to-control path and are introduced by changing task objects, commands, objectives, and curricula.} We denote such a definition by $\mathcal{T}=(\mathcal{O},\mathbf{c}_t,\mathbf{y}_t,r_t,\mathcal{K})$, where $\mathcal{O}$ is the set of task objects, $\mathbf{c}_t$ is the optional command, $\mathbf{y}_t$ is the task geometry supervised through Sec.~\ref{sec:visibility-aux}, $r_t$ defines the contact objective, and $\mathcal{K}$ is the curriculum. \AuJK{We use goal-directed shooting and directional dribbling as representative instances to verify the framework across complementary contact regimes; following details illustrate how the shared framework is instantiated for these examples.}

%% file: sections/method/shooting.tex
Shooting is defined by the ball--goal geometry. Let $\mathbf{b}_t$ and $\mathbf{g}_t$ denote the ball position and goal center in the head-yaw-aligned frame, and let $\mathbf{d}^{bg}_t=\frac{\mathbf{g}_t-\mathbf{b}_t}{\|\mathbf{g}_t-\mathbf{b}_t\|_2+\epsilon}$ be the normalized ball-to-goal direction. A behind-ball target can then be written as $\mathbf{x}^{\mathrm{behind}}_t=\mathbf{b}_t-\rho\mathbf{d}^{bg}_t$, which turns chasing, alignment, and contact into one geometric target family. In this task, the object set is $\mathcal{O}_{\mathrm{shoot}}=\{\mathrm{ball}\,,\,\mathrm{goal}\}$, the task geometry is $\mathbf{y}_t^{\mathrm{shoot}}=(\mathbf{b}_t,\mathbf{g}_t)$, and the objective is to strike the ball so that its exit motion moves toward the goal. The reward is organized around run-up, contact, terminal success, visual maintenance, and safety regularization, while the curriculum gradually broadens the initial ball placement and approach difficulty. Active vision is maintained through the head DoFs: the head keeps the ball within view, and the body gradually follows the viewing direction.

%% file: sections/method/dribbling.tex
Dribbling is defined by a direction command and sustained ball regulation.
Let $c_t\in\mathcal{C}$ be the commanded direction angle, where
$\mathcal{C}=\{k\pi/6:k=-6,\ldots,5\}$ is a 12-way ring uniformly
spaced over $360^\circ$, and
$\mathbf{d}^{\mathrm{cmd}}_t=[\cos c_t,\sin c_t]$
the corresponding horizontal direction.
The task object set is
$\mathcal{O}_{\mathrm{dribble}}=\{\mathrm{ball}\}$,
and the auxiliary geometry is the 3-D ball center
$\mathbf{y}_t^{\mathrm{ball}}=\mathbf{b}_t$.
The commanded direction $\mathbf{d}^{\mathrm{cmd}}_t$ and the
robot--ball horizontal distance $d_t$ are task variables used to define
the dribbling objective rather than outputs of the auxiliary geometry head.
Unlike shooting, the objective is not a single high-speed strike; the policy
must keep the ball within a controllable band and generate progress along
$\mathbf{d}^{\mathrm{cmd}}_t$.
We express the controllable-band condition by
$\chi_t^{\mathrm{band}}=
\mathbb{I}\!\left(
|d_t-d_0|<\Delta_d,\,
v_{lo}<\|\mathbf{v}_t^b\|<v_{hi},\,
\mathbf{v}_t^b\cdot\mathbf{d}^{\mathrm{cmd}}_t>0
\right)$.
The reward covers commanded progress, re-approach, heading, visual maintenance,
and out-of-band penalties.
Continuous dribbling further uses contact-intent-conditioned AMP and phase masks
so approach, touch, follow-through, and recovery emphasize different motion states.

%% file: sections/method/pipeline.tex
The deployment policy uses a single-frame head-mounted depth image, a five-step proprioceptive history, and task variables to output 25-DoF joint PD targets. Since this depth image is the main sim-to-real interface, we align simulation and deployment before it enters the actor. As shown in Fig.~\ref{fig:pipeline}, a raw frame $\mathbf{D}_t$ is processed to $\tilde{\mathbf{D}}_t$. During training, the aligned frame is further perturbed to reduce dependence on clean depth and serve as domain randomization for sim-to-real transfer~\cite{tobin2017domain,peng2018simtoreal}. \AuJK{Both simulated and real depth frames are cropped to $168\times80$, clipped to 0.3--5.0\,m, normalized, and zero-filled at invalid pixels. Policies are trained in Isaac Lab for 20,000 PPO iterations with AMP and a history-informed latent estimator (HIM)~\cite{long2024hybrid}, then exported to ONNX. On the robot, the actor runs at 50\,Hz, the low-level PD controller at 200\,Hz, and the depth stream at 15\,Hz; no external localization, object detector, state estimator, or runtime soccer planner is used.}

%% file: sections/experiments/overview.tex
We evaluate whether \textbf{DAVIS} supports both simulated and real-world soccer contact skills, as illustrated in Fig.~\ref{fig:qualitative-soccer}. And we answer the following questions: \textbf{Q1.} Can a depth-only end-to-end policy perform soccer contact skills? \textbf{Q2.} Which training structures are necessary? \textbf{Q3.} Does the learned policy transfer to the real Noetix E1 robot?

%% file: sections/experiments/simulation.tex
To answer Q1, we extensively evaluate \textbf{DAVIS} in IsaacLab simulator on two representative skill categories in soccer: shooting and dribbling.

Shooting and directional dribbling are trained as separate task-specific checkpoints that share the same depth-to-control interface; they are not a single multi-skill policy. At deployment, the shooting actor receives no operator command, while the dribbling actor receives only the intended direction from a joystick, quantized to the same 12-way ring defined in Sec.~\ref{sec:dribbling-instantiation}.

\textbf{Shooting}: The robot is required to perform shooting with randomly initialized robot or ball positions. In the penalty kick task, the robot starts within an angle of $[-100^\circ,100^\circ]$ behind the ball and distances of $2$--$3.5\,\mathrm{m}$. In the free kick task, the robot is initialized $5\,\mathrm{m}$ in front of the goal, and is required to kick the ball within the region $x \in [0,4]\,\mathrm{m}$ and $y \in [-3,+3]\,\mathrm{m}$. The difficulty level categories are shown in Fig.~\ref{fig:shooting_results}. We record the success rate of the task. Experimental results in Table~\ref{tab:quantitative-results} demonstrate strong positional generalization capability. By analyzing the kinematic pattern during the shooting process in Fig.~\ref{fig:shooting_results}, we observe a smooth gait and consistent speed pattern.

\textbf{Dribbling}: The robot is required to complete a repeated-S slalom task. $N$ is the number of poles, and $\alpha$ is the turning angle of instruction. A task is considered successful if the robot passes through all waypoints in the predefined order. To ensure consistency, we use a fixed instruction sequence for all experiments. The policy achieves an average success rate of 0.65 across 900 trials. Experimental results in Table~\ref{tab:quantitative-results} demonstrate that more than half of the task configurations obtain success rates above 0.65, showing robustness in long trajectory and extreme angles. The visualization in Fig.~\ref{fig:dribble-sim} shows the policy's dynamic behavior.   

\AuJK{\textbf{Head-control comparison}:} We compare end-to-end learned head control with a heuristic that gazes at the predicted ball (Table~\ref{tab:unified-ablation}, last row). On static objects such as penalty kicks, the heuristic is comparable to learned active vision. On highly dynamic dribbling, the heuristic fails while the learned controller remains effective. The gap confirms that head motion must be learned jointly with locomotion and contact, rather than as a detached look-at-ball module, motivating active vision within the end-to-end policy.

\AuJK{\textbf{Comparison with an external protocol}: Dribble Master~\cite{wang2025dribblemaster} is the closest published system to our setting. Both use full-size humanoids with 2-DoF active heads and onboard sensing only; however, its policy consumes detector-derived ball coordinates, whereas DAVIS consumes a depth image. We evaluate DAVIS on its two core dribbling tasks with 50 trials per condition. At the $1$\,m tolerance, DAVIS achieves target-reaching and obstacle-avoidance success rates of $0.88$ and $0.96$, compared with the reported Dribble Master rates of $0.87$ and $0.93$.}

%% file: tables/ablation/unified_ablation.tex
\begin{table*}[t]
  \centering
  \scriptsize
  \setlength{\tabcolsep}{2.2pt}

  \newlength{\firstcolwidth}
  \setlength{\firstcolwidth}{5.2cm}

  \newlength{\mytabcolwidth}
  \setlength{\mytabcolwidth}{%
    \dimexpr(\textwidth - \firstcolwidth - 18\tabcolsep)/9\relax
  }

  \newlength{\mytabtotalwidth}
  \setlength{\mytabtotalwidth}{%
    \dimexpr\firstcolwidth + 9\mytabcolwidth + 18\tabcolsep\relax
  }

  \newlength{\mygraywidth}
  \setlength{\mygraywidth}{\mytabtotalwidth}

  \newcolumntype{A}{>{\centering\arraybackslash}m{\mytabcolwidth}}
  \newcolumntype{L}{>{\raggedright\arraybackslash}m{\firstcolwidth}}

  \renewcommand{\arraystretch}{1.0}

  \newcommand{\rowstrut}{\rule{0pt}{2.4ex}\rule[-1.0ex]{0pt}{0ex}}

  \newcommand{\ablationheader}[1]{%
    \noalign{\global\aboverulesep=0pt \global\belowrulesep=0pt}%
    \hline
    \rowcolor{gray!15}%
    \multicolumn{10}{>{\raggedright\arraybackslash}m{\mygraywidth}}{%
      \rule{0pt}{2.4ex}\textbf{#1}\rule[-1.0ex]{0pt}{0ex}%
    }%
    \\[0pt]
    \hline
    \noalign{\global\aboverulesep=0.65ex \global\belowrulesep=0.65ex}%
  }

  \caption{Simulation ablations and comparison. 
  (a) Ablations. Penalty-kick and free-kick results use over 900 shooting trajectories, 
  and repeated-S slalom uses over 100 dribbling trajectories. 
  Metrics are success rate (SR), body-to-goal angular error ($E_{\mathrm{ang}}$), 
  successful ball-contact ratio ($r_{\mathrm{contact}}$), 
  velocity error ($E_{\mathrm{vel}}$), and time in the control band ($r_{\mathrm{bib}}$). 
  The full method provides the best overall balance across tasks and metrics. 
  (b) Comparison between our end-to-end learned head control (Active vision) and a heuristic head controller (Heuristic) that gazes at the predicted ball. The heuristic performs comparably to active vision in static-object tasks (e.g., penalty kicks), while active vision shows a clear advantage in highly dynamic dribbling scenarios.}
  \vspace{0.5em}
  \label{tab:unified-ablation}

  \begin{tabular}{@{}L*{9}{A}@{}}
    \toprule
    \multirow{2}{*}{\rowstrut\raisebox{-0.7ex}{Method}}
    & \multicolumn{3}{c}{Penalty kick}
    & \multicolumn{3}{c}{Free kick}
    & \multicolumn{3}{c}{Repeated-S slalom} \\
    \cmidrule(lr){2-4}\cmidrule(lr){5-7}\cmidrule(lr){8-10}
    & SR $\uparrow$ & $E_{\mathrm{ang}}\downarrow$ & $r_{\mathrm{contact}}\uparrow$
    & SR $\uparrow$ & $E_{\mathrm{ang}}\downarrow$ & $r_{\mathrm{contact}}\uparrow$
    & SR $\uparrow$ & $E_{\mathrm{vel}}\downarrow$ & $r_{\mathrm{bib}}\uparrow$ \\
    \midrule

    \raisebox{0.35ex}{\textbf{DAVIS}}\rowstrut
    & 0.84 & \textbf{23.29} & \textbf{99.75}
    & \textbf{0.80} & \textbf{30.18} & \textbf{99.27}
    & \textbf{0.70} & 0.79 & \textbf{0.74} \\

    \ablationheader{(a) Ablation}
    \raisebox{-0.4ex}{DAVIS-w/o-curriculum}\rowstrut
    & \raisebox{-0.65ex}{0.62} & \raisebox{-0.65ex}{27.40} & \raisebox{-0.65ex}{32.53}
    & \raisebox{-0.65ex}{0.55} & \raisebox{-0.65ex}{34.42} & \raisebox{-0.65ex}{87.67}
    & \raisebox{-0.65ex}{0.51} & \raisebox{-0.65ex}{\textbf{0.77}} & \raisebox{-0.65ex}{0.69} \\
    \midrule

    DAVIS-auxiliary head-w/o-anneal\rowstrut
    & \raisebox{-0.65ex}{0.32} & \raisebox{-0.65ex}{54.79} & \raisebox{-0.65ex}{84.87}
    & \raisebox{-0.65ex}{0.22} & \raisebox{-0.65ex}{76.20} & \raisebox{-0.65ex}{87.92}
    & \raisebox{-0.65ex}{0.31} & \raisebox{-0.65ex}{0.83} & \raisebox{-0.65ex}{0.66} \\
    DAVIS-auxiliary head-w/o-geometry\rowstrut
    & \raisebox{-0.65ex}{0.20} & \raisebox{-0.65ex}{53.53} & \raisebox{-0.65ex}{87.33}
    & \raisebox{-0.65ex}{0.15} & \raisebox{-0.65ex}{69.47} & \raisebox{-0.65ex}{78.13}
    & \raisebox{-0.65ex}{0.34} & \raisebox{-0.65ex}{0.91} & \raisebox{-0.65ex}{0.54} \\
    DAVIS-auxiliary head-w/o-supervision\rowstrut
    & \raisebox{-0.65ex}{0.13} & \raisebox{-0.65ex}{81.87} & \raisebox{-0.65ex}{93.33}
    & \raisebox{-0.65ex}{0.14} & \raisebox{-0.65ex}{93.66} & \raisebox{-0.65ex}{91.67}
    & \raisebox{-0.65ex}{0.30} & \raisebox{-0.65ex}{0.95} & \raisebox{-0.65ex}{0.48} \\
    DAVIS-auxiliary head-w/o-visibility gate\rowstrut
    & \raisebox{-0.65ex}{0.34} & \raisebox{-0.65ex}{51.97} & \raisebox{-0.65ex}{91.33}
    & \raisebox{-0.65ex}{0.29} & \raisebox{-0.65ex}{68.44} & \raisebox{-0.65ex}{92.08}
    & \raisebox{-0.65ex}{0.27} & \raisebox{-0.65ex}{1.06} & \raisebox{-0.65ex}{0.64} \\
    DAVIS-w/o-auxiliary head\rowstrut
    & \raisebox{-0.65ex}{0} & \raisebox{-0.65ex}{--} & \raisebox{-0.65ex}{0}
    & \raisebox{-0.65ex}{0} & \raisebox{-0.65ex}{--} & \raisebox{-0.65ex}{0}
    & \raisebox{-0.65ex}{0} & \raisebox{-0.65ex}{1.45} & \raisebox{-0.65ex}{0.05} \\
    \midrule

    \raisebox{0.25ex}{DAVIS-w/o-active vision}\rowstrut
    & 0 & -- & 0 & 0.36 & 97.81 & 1.71 & 0 & 1.25 & 0.03 \\

    \ablationheader{(b) Comparison}
    \raisebox{-0.4ex}{DAVIS-Head controller}\rowstrut
    & \raisebox{-0.75ex}{\textbf{0.88}} & \raisebox{-0.75ex}{26.66} & \raisebox{-0.75ex}{93.67}
    & \raisebox{-0.75ex}{0.76} & \raisebox{-0.75ex}{37.96} & \raisebox{-0.75ex}{88.13}
    & \raisebox{-0.75ex}{0.0} & \raisebox{-0.75ex}{0.99} & \raisebox{-0.75ex}{0.13} \\
    \bottomrule
  \end{tabular}
  \vspace{-2.5em}

\end{table*}

%% file: sections/experiments/ablation.tex
\AuJK{To verify the necessity of key structures and answer Q2, we conduct ablation studies on three tasks in simulation and record different metrics in Table~\ref{tab:unified-ablation}.} All methods share the same training parameters and evaluation settings.

\textbf{Ablation on curriculum.} We remove the ramp-up and scaling curriculum. All reward terms are activated with full weights from the beginning. Without the progressive curriculum, the policy requires more training iterations to converge, and performance is significantly worse than baseline. 

\textbf{Ablation on auxiliary head.} \AuJK{Our policy incorporates an object position prediction head.} To validate effectiveness, we mask the predicted positions fed into the actor, while retaining the network architecture and input dimension. \AuJK{We disable the aux loss and stop backpropagation.} As expected, without the explicit predictions, the model struggles to localize objects to make contact and success drops to zero. To further isolate each switch of the full auxiliary head, we disaggregate this into four finer variants: w/o anneal keeps $\beta_k=1$ throughout (always predicted auxiliary features; no GT$\to$pred schedule); w/o geometry zeros the actor's auxiliary input while keeping auxiliary losses; w/o supervision drops the goal/ball/visibility losses but still feeds predicted features; w/o vis. gate skips visibility$\times$geometry masking. 

\textbf{Ablation on active vision.} Active vision is another key design.\ZXX{ Since its components all act through the same two head-joint control dimensions and cannot be ablated independently, } \AuJK{We fix the 2-DoF head joints, replace the original head-based rewards with body alignment rewards, and adjust the curriculum.} After fixing the head joints, the ball frequently moves out of the FOV or enters blind spots, causing collapse. Another notable observation is that the robot is unable to locate the gate. The performance drops nearly to zero, with a high likelihood of falling.

%% file: sections/experiments/real_world.tex
To answer Q3, we deploy our policies on the Noetix E1 humanoid robot equipped with a ZED2i depth camera mounted on a 2-DoF head. For the shooting experiments, we evaluate 12 predefined positions in one penalty-kick protocol; the Easy, Medium, and Hard settings use progressively more distant and more oblique positions, with 38, 49, and 60 repeated trials, respectively. For dribbling experiments, we design three tasks including reaching target position, obstacle traversal, and slalom dribbling. To better reflect real-world soccer scenarios, all experiments are conducted on a grass surface. Safety protection uses a tether or staff member; emergency-stop and projected-gravity fall triggers terminate unsafe rollouts, which are restarted from their initial conditions. If the ball is lost without an active recovery behavior, the contact policy exits to the default walk--run policy. Fig.~\ref{fig:teaser} and Fig.~\ref{fig:qualitative-soccer} show that our policy can perform dynamic soccer tasks with precise actions. Quantitative results in Table~\ref{tab:quantitative-results} demonstrate that our policy maintains a high success rate in real deployment. \AuJK{Shooting records 26/38, 28/49, and 33/60 successes for the Easy, Medium, and Hard settings, respectively; a success requires the ball to cross the goal line in one kick. Dribbling records 22/28, 19/28, and 17/28 successes for target reaching, obstacle traversal, and slalom, respectively; a successful dribbling trial requires the robot to remain upright throughout execution and both the robot and ball to finish within 1\,m of the target. All dribbling trials were conducted by the same operator using a joystick controller.}

%% file: sections/conclusion.tex
We presented a depth-only end-to-end active-vision framework \textbf{DAVIS} for humanoid soccer skills. The main idea is simple: keep the deployment interface minimal, but place enough structure inside training so that the policy can learn soccer-specific geometry, visibility, and whole-body contact behavior from depth and proprioception alone.

\AuJK{Within the shared framework, goal-directed shooting and directional dribbling serve as representative instances of one-shot striking and sustained ball control, retaining the observation-to-control interface while changing task-specific objects, commands, rewards, curricula, and motion-prior conditioning. Across simulation, ablations, and the complete Noetix E1 deployment pipeline, real-robot execution confirms that this depth-only interface transfers across contact regimes; together with additional skill demonstrations, these results support DAVIS as an extensible framework for humanoid soccer skills through task-specific training.}

%% file: sections/limitations.tex
Two limitations remain. The first is perceptual: depth-only sensing keeps the interface compact but ties the policy to a single head-mounted camera, inheriting its occlusions, dropouts, and imperfectly simulated depth---a residual sim-to-real gap that limits real-world accuracy. The second sits upstream: the task-definition interface unifies shooting and dribbling, yet each skill needs hand-tuned rewards, curricula, and contact conditioning, leaving the pipeline a recipe, not an automatic one.

Closing the perceptual gap is largely engineering---sensor-noise modeling and real-world fine-tuning. Automating skill construction is the more open problem, where language-conditioned specification and automatic curricula could generate new behaviors on the same interface.

%% file: appendix/appendix.tex

\section{Robot System and Deployment Interface}
\label{app:robot-system}
\input{appendix/appendix_A.tex}

\section{Evaluation Protocols and Metric Definitions}
\label{app:protocols}
\input{appendix/appendix_B.tex}

\section{Observation Space, Optimization, and Auxiliary Supervision}
\label{app:policy-architecture}
\input{appendix/appendix_C.tex}

\section{Task Definitions and Reward Details}
\label{app:task-details}
\subsection{Goal-Directed Shooting}
\label{app:shooting-details}
\input{appendix/appendix_D.1Goal-Directed_Shooting.tex}

\subsection{Directional Dribbling}
\label{app:dribbling-details}
\input{appendix/appendix_D.2_Directional_Dribbling.tex}

\section{Domain Randomization and Sim-to-Real Training}
\label{app:sim2real}
\input{appendix/appendix_E_sim2real.tex}

\section{Additional Skills and Results}
\label{app:additional-results}
\input{appendix/additional_results.tex}

%% file: appendix/appendix_A.tex
For reproducibility, policies are trained in Isaac Lab (built on Isaac Sim 5.0.0) on eight NVIDIA RTX 4090D GPUs for 20,000 PPO iterations before ONNX export. The 25-dimensional action vector is preserved across training, export, and deployment; hardware rates, sensor inputs, and excluded runtime modules are specified in Sec.~\ref{sec:real-world}.

%% file: appendix/appendix_B.tex
Table~\ref{tab:metric-definitions} defines the metrics used by the simulation,
ablation, and real-world result tables. The main paper gives the task-level
success definitions; this appendix records the protocol details that expand
those definitions for each evaluation setting.

\begin{table}[htbp]
\centering
\caption{Metric definitions used throughout the evaluation.}
\label{tab:metric-definitions}
\begin{tabularx}{\linewidth}{>{\raggedright\arraybackslash}m{0.25\linewidth}>{\raggedright\arraybackslash}X}
\toprule
Metric & Definition \tabularnewline
\midrule
Success rate (SR) & The main paper defines task-level success: shooting requires the ball to cross the goal line in one kick, while dribbling requires the robot to remain upright and both the robot and ball to finish within $1$\,m of the target. This appendix expands that definition for evaluation protocols: simulated dribbling additionally requires visiting all waypoints in order (Sec.~\ref{sec:simulation}), whereas the real-world protocol applies the same endpoint and uprightness criteria to each trial. \tabularnewline
Valid contact ratio ($r_{\mathrm{contact}}$) & Fraction of shooting trials with a valid foot--ball contact and goal-directed ball exit. \tabularnewline
Angular error ($E_{\mathrm{ang}}$) & Heading error between the robot body and the robot-to-goal direction. \tabularnewline
Velocity error ($E_{\mathrm{vel}}$) & Mean error between commanded and actual ball velocity during dribbling. \tabularnewline
Ball-in-band ratio ($R_{\mathrm{bib}}$) & Fraction of engaged dribbling time in which the ball stays inside the controllable band. \tabularnewline
Turn bias ($B_{\mathrm{turn}}$) & $|\bar\theta_{\mathrm{achieved}}-\theta_{\mathrm{cmd}}|/\theta_{\mathrm{cmd}}$ over scored rounds under the Dribble Master turning protocol; the statistic behind its reported turning error. \tabularnewline
Reach rate ($\mathrm{RR}_{\tau}$) & Fraction of rounds whose minimum ball--goal distance falls below tolerance $\tau$ under the Dribble Master reaching protocol. \tabularnewline
\bottomrule
\end{tabularx}
\end{table}

Each main-paper result table follows either Table~\ref{tab:metric-definitions}
or the task-specific protocol described in this appendix. When a metric is
computed differently for simulation and real-world deployment, the difference is
reported with the corresponding result table.

\subsection{Directional Shooting Simulation 
Protocol}\label{app:shooting-protocol}
\begin{figure}[htbp]
\centering
\includegraphics[width=\linewidth]{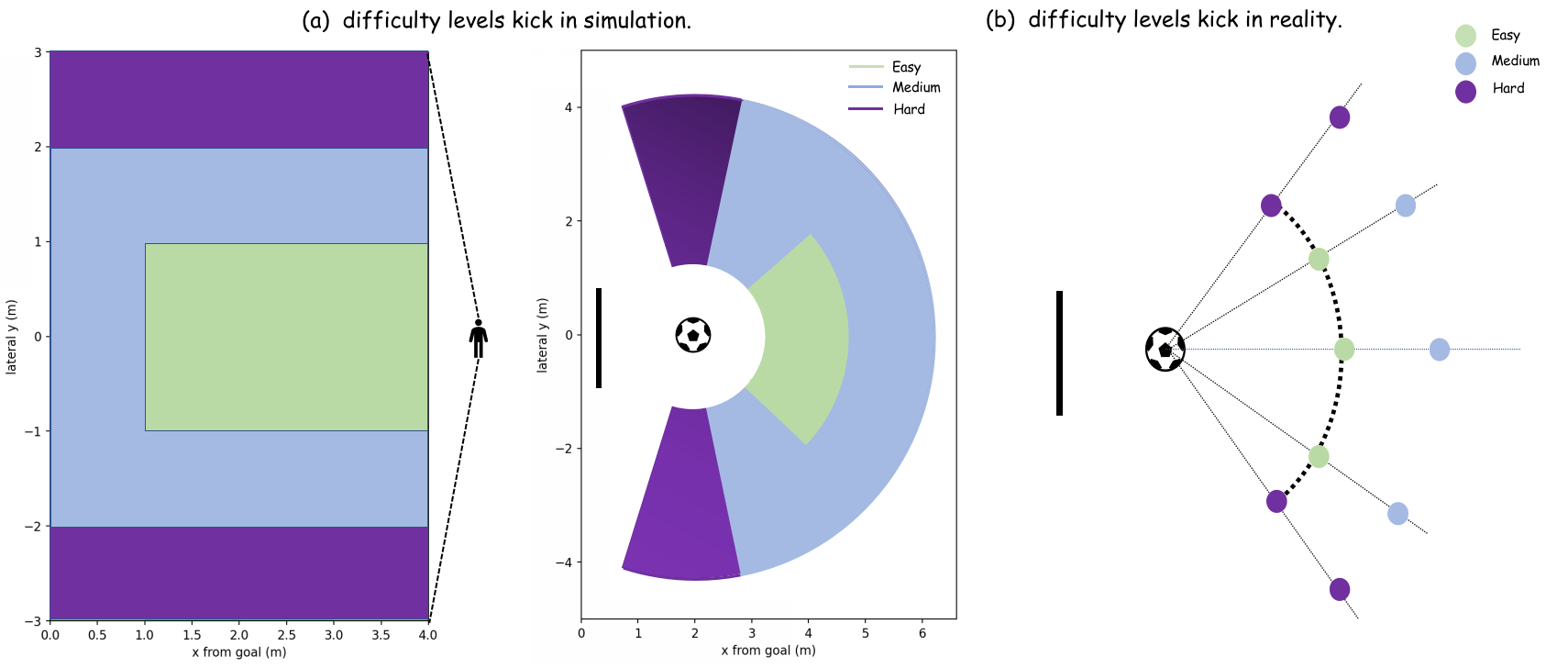}
\caption{
Multi-position Shooting Simulation and Real Machine Difficulty Level Classification
}
\label{fig:difficulty-levels}
\end{figure}

The shooting simulation results in Table~\ref{tab:quantitative-results} and Fig.~\ref{fig:shooting_results} are evaluated on two tasks: penalty kick and free kick. In the free-kick task, the robot starts from a fixed position 5\,m in front of the goal center and faces the goal, while the ball is randomly placed within $x \in [0,4]\,\mathrm{m}$ and $y \in [-3,+3]\,\mathrm{m}$. In the penalty-kick task, the ball is fixed 1.5\,m in front of the goal center, while the robot is initialized behind the ball within an angular range of $[-100^\circ,100^\circ]$ and a distance range of $2$--$3.5\,\mathrm{m}$, facing the ball. The difficulty level diagram of the shooting task is shown as Figure~\ref{fig:difficulty-levels}.

To evaluate spatial generalization, we use a stratified random initialization scheme. The penalty-kick sector is divided into $2 \times 15$ regions, and the free-kick ball-placement area is discretized into $1\,\mathrm{m} \times 1\,\mathrm{m}$ grid cells. We sample the same number of trials from each region to reduce sampling bias and ensure uniform coverage.

We report SR, $E_{\mathrm{ang}}$, and $r_{\mathrm{contact}}$ following the definitions in Table~\ref{tab:metric-definitions}. The angular error is computed as
\begin{equation}
E_{ang}
=
\arccos\left(
\frac{\mathbf{v}_1^\top \mathbf{v}_2}
{\|\mathbf{v}_1\|\|\mathbf{v}_2\|}
\right)
\end{equation}
where $\mathbf{v}_1$ is the robot body-heading vector and $\mathbf{v}_2$ points from the robot base to the goal center. All shooting simulation evaluations use the same policy checkpoint, IsaacLab environment configuration, 5\,m depth range, and depth preprocessing pipeline described in Sec.~\ref{sec:training-deployment}.

\subsection{Directional Dribbling Simulation Protocol}\label{app:dribble-protocol}

The dribbling simulation results in Table~\ref{tab:quantitative-results} are evaluated by replaying a fixed policy checkpoint in the same IsaacLab environment and control configuration used for training. The benchmark uses a parametrized repeated-S slalom fixture (Table~\ref{tab:dribble-eval-fixture}) over $N\in{3,5,7}$ trajectory turns and $\alpha\in{60,90,120}^{\circ}$ turning angles, resulting in nine evaluation tasks and 900 trials in total. The ball is initialized 0.6\,m in front of the robot, while the robot pose and AMP reference-state reset vary across rounds.

\begin{table*}[t]
\centering
\caption{Directional dribbling simulation fixture.}
\label{tab:dribble-eval-fixture}
\begin{tabularx}{\linewidth}{>{\centering\arraybackslash}m{0.18\linewidth}>{\raggedright\arraybackslash}X>{\centering\arraybackslash}m{0.29\linewidth}}
\toprule
Field & Definition & Value \tabularnewline
\midrule
Task grid & Repeated-S slalom with trajectory turns $N$ and turn angle $\alpha$ & $N\in{3,5,7}$, $\alpha\in{60,90,120}^{\circ}$ \tabularnewline
Waypoints & $N{+}2$ turning waypoints plus $N{+}1$ midline gate points & $2N{+}3$ total \tabularnewline
Geometry & Fixed arm length for waypoint generation & $L=4.0$\,m \tabularnewline
Command & Waypoint attraction and pole repulsion, snapped to 12 direction bins & $k_{\mathrm{att}}=1.0$, $k_{\mathrm{rep}}=0.6$, $d_0=1.2$\,m \tabularnewline
Ball spawn & Fixed point directly in front of the robot & 0.6\,m \tabularnewline
Engagement & Metrics start after first ball contact & threshold 0.15\,m/s; timeout 5\,s \tabularnewline
Waypoint arrival & Radius for advancing to the next waypoint & 0.75\,m \tabularnewline
Segment timer & Time budget refreshed after each waypoint arrival & 5\,s \tabularnewline
Failure cases & Fall, timeout, or ball leaving the controllable band & --- \tabularnewline
\bottomrule
\end{tabularx}
\end{table*}

Following Table~\ref{tab:metric-definitions}, we report SR, $E_{\mathrm{vel}}$, and $R_{\mathrm{bib}}$ for dribbling. All metrics are computed over the post-engagement window, i.e., after the robot first takes over the ball, so the initial approach phase is excluded. Specifically, $E_{\mathrm{vel}}$ is the mean L2 error between the commanded ball velocity and the measured ball velocity over the engaged steps, where the commanded velocity points along the current dribbling command with a peak speed of $1.0\,\mathrm{m/s}$. $R_{\mathrm{bib}}$ is accumulated over the same engaged window and is undefined for rounds that never enter the controllable band.

The in-band predicate follows the deployed termination settings: the robot--ball distance must lie within 0.10\,m of the 0.42\,m target, the ball speed must lie in $[0.20,1.50]$\,m/s, and the ball velocity must not point against the command direction. A round fails if the ball remains out of band for more than 3.5\,s or exceeds the 0.80\,m far-field scope. For fair aggregation, every policy is evaluated with the same checkpoint selection rule, fixed ball spawn, slalom grid, band thresholds, and reset distribution; only the policy weights differ across ablations.
\begin{figure*}[t]
\centering
\includegraphics[width=\linewidth]{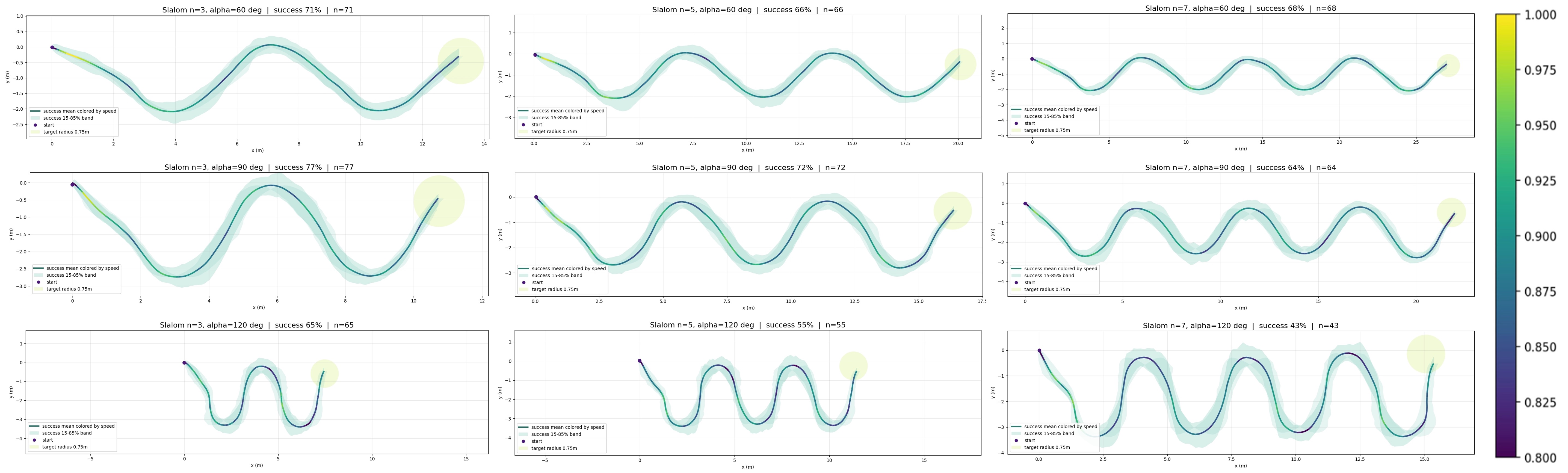}
\caption{
Trajectory visualization for the repeated-S slalom dribbling benchmark.
Each subplot corresponds to one task configuration $(N,\alpha)$, where $N$
is the number of turns and $\alpha$ is the turning angle. The colored curve
shows the mean successful ball trajectory with speed-based coloring, and the
shaded region indicates the 15--85\% success band. The start point and target
region are marked for reference. The title of each subplot reports the success
rate and the number of successful rounds.
}
\label{fig:dribble-traj}
\end{figure*}

\paragraph{Trajectory visualization.}
Figure~\ref{fig:dribble-traj} visualizes the repeated-S slalom trajectories. Each panel corresponds to a task configuration $(N,\alpha)$ and shows engagement-anchored paths in a local forward-left frame, the mean successful trajectory, its success band, ball-speed coloring, the start point, and the target region. The figure uses the same checkpoint and evaluation protocol as the metric tables.

\subsection{Dribble Master Protocol Reproduction}\label{app:dm-protocol}

We reproduce the dribbling evaluation of Dribble Master~\cite{wang2025dribblemaster} on DAVIS with the checkpoint and simulator of Table~\ref{tab:quantitative-results}. Table~\ref{tab:dm-protocol} aligns the two setups and Tables~\ref{tab:dm-results} and~\ref{tab:dm-turning-results} report the full results; the main-text figures are the $d{=}3$\,m reaching and $d{=}5$\,m obstacle cells at the $1$\,m tolerance. Turning is reported here only. The achieved turn is the angle between the ball's mean travel directions over $0.6$\,s before the command switch and over a $1.5$\,s window starting $2.0$\,s after it; its bias of the mean, the statistic behind their published percentages, averages $7.1\%$ over the eight conditions against their $2.2$--$3.2\%$. The gap is consistent with the command interface---they track a ball-velocity vector, whereas DAVIS commands direction only and leaves lateral drift inside the controllable band uncorrected---rather than with perception, and the off-grid $45^\circ$ conditions track no worse than the neighbouring on-grid $30^\circ/60^\circ$ controls.
\input{tables/appendix/dm_protocol.tex}

\input{tables/appendix/dm_results.tex}

\subsection{Ablation Experiment Protocol} \label{ablation_protocol}

For all variant in ablation experiments, We follow the identical test settings, evaluation protocols and metrics mentioned in Appendix \ref{app:shooting-protocol} and \ref{app:dribble-protocol}. Meanwhile, we adopt identical learning hyperparameters, training steps and parallel environments as those used in simulation experiments to guarantee consistent experimental variables. For safety concerns, no ablation studies are conducted on real-world robots. 

\noindent\textbf{Ablation on curriculum.}
The ramp-up of head and body rewards groups is removed.
From step 0 onward, body and head reward scale are fixed to 1.
No staged scaling is applied to ball-chasing and gaze-related terms.
All task rewards and ball-chasing components take full effect from the beginning of training,
eliminating the staged curriculum of ``movement first, head activation later''.

\noindent\textbf{Ablation on auxiliary head.}
In the full architecture, depth maps are encoded into latent features via a CNN network.
Three prediction heads (goal, ball, visibility) then generate geometric and visibility features, which are concatenated to the actor input.
An auxiliary loss is constructed using privileged ground truth to update the encoder and prediction heads.
In this ablation, we retain the network structure of the prediction heads and the dimension of the auxiliary input to the actor,
but mask all auxiliary features fed into the actor to 0, disable the auxiliary loss, and cut off its backpropagation.
This renders the prediction head fully non-functional during both training and inference without changing the overall model structure.

\begin{figure}[htbp]
  \centering
  \includegraphics[width=\linewidth]{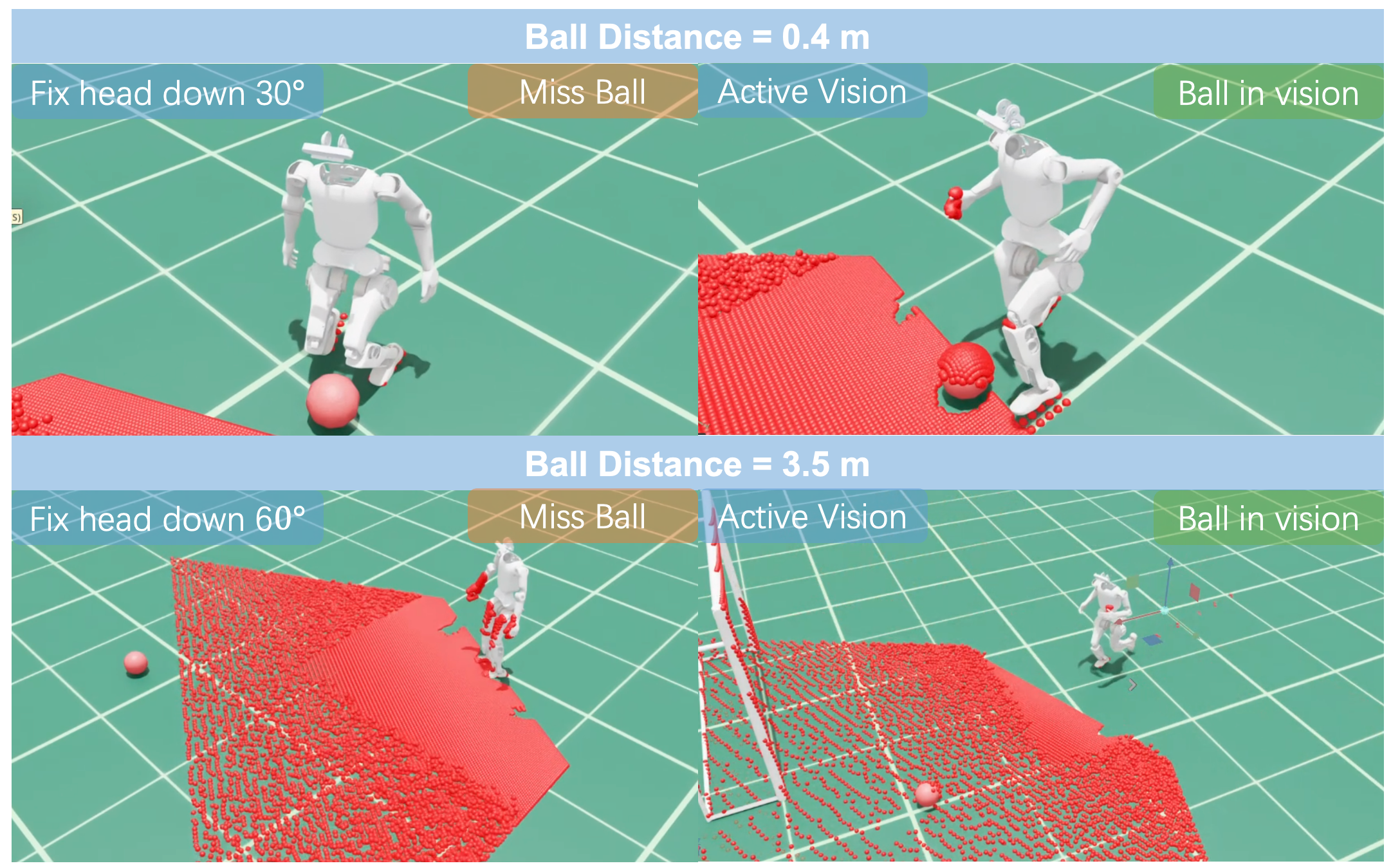}
  \caption{Ablation on active vision.}
  \label{fig:head_ablation}
  
\end{figure}

\noindent\textbf{Ablation active vision.}
The reward structure for ball chasing remains unchanged,
but the control objective is switched from ``where to orient the head'' to ``where to rotate the body''.
The two original terms relying on head joint angles are replaced with body yaw/pitch to align with the ball and target line.
Accordingly, the curriculum learning is adjusted from head reward scale to body reward scale,
allowing the constraint to be gradually activated together with the body-following reward,
instead of using an independent head-oriented curriculum. Figure~\ref{fig:head_ablation} visualizes this active-vision ablation.

\subsection{Real World Experiment Protocol}\label{app:real-protocol}

\begin{figure}[htbp]
\centering
\includegraphics[width=\linewidth]{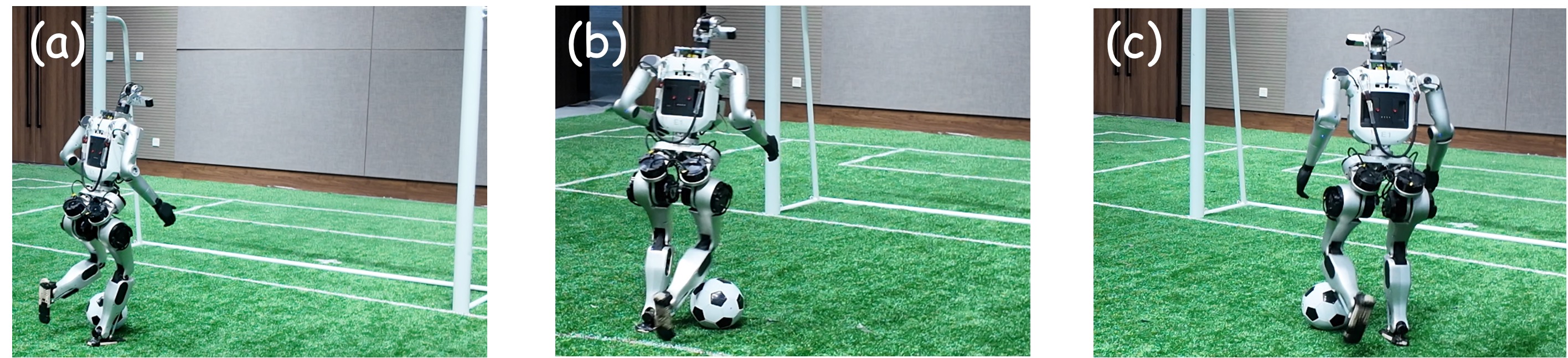}
\caption{
Multi-directional shooting real-world example
}
\label{fig:shoot}
\end{figure}

\begin{figure}[htbp]
  \centering
  \includegraphics[width=\linewidth]{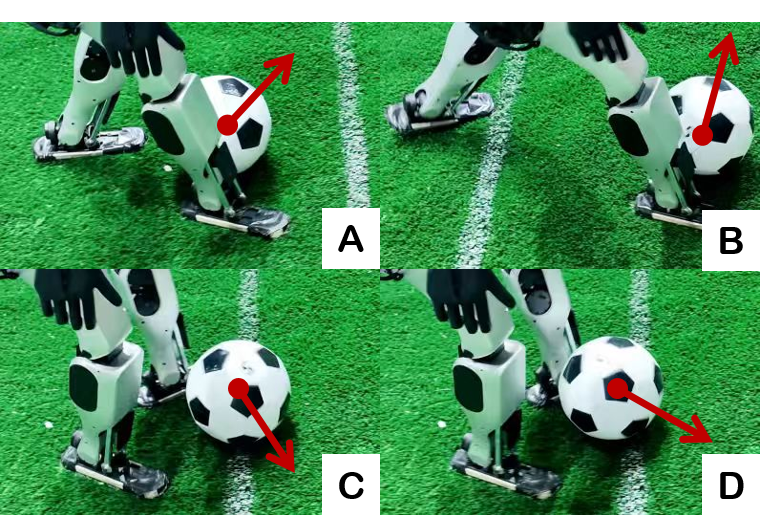}
  \caption{Foot snapshot in dribbling experiment.}
  \label{fig:dribble_foot}
  
\end{figure}

Figures~\ref{fig:shoot} and~\ref{fig:dribble_foot} show real-world shooting and dribbling examples, respectively.

All real-world experiments follow the settings detailed in Appendix~\ref{app:robot-system}, use safety protection on a grass field, and report the task success rate defined in Table~\ref{tab:metric-definitions}.

\textbf{Real World Shooting Experiment.} For the real-world shooting experiments showed in Table~\ref{tab:quantitative-results}, we adopt the penalty-kick setting. The ball is placed \(1.5\,\mathrm{m}\) in front of the goal, and the robot performs penalty kicks from different initial positions. The goal dimensions match those in simulation: half-width \(1.20\,\mathrm{m}\), height \(1.80\,\mathrm{m}\), and depth \(1.00\,\mathrm{m}\).

\begin{table}[t]
\centering
\caption{Target point configurations and difficulty levels used in the real-world shooting evaluation.}
\label{tab:real_world_targets}
\begin{tabularx}{\linewidth}{@{}>{\centering\arraybackslash}m{0.28\linewidth}>{\centering\arraybackslash}m{0.20\linewidth}>{\centering\arraybackslash}X@{}}
\toprule
Difficulty & Distance (m) & Target Angles \\
\midrule
Easy   & 2.5 & $-30^\circ$, $-5^\circ$, $+5^\circ$, $+30^\circ$ \\
Medium & 3.0      & $-30^\circ$, $-5^\circ$, $+5^\circ$, $+30^\circ$ \\
Hard   & 2.5--3.0      & $-45^\circ$, $+45^\circ$ \\
\bottomrule
\end{tabularx}
\end{table}

To ensure experimental consistency, we evaluate the policy from 12 predefined positions within the test area. The Easy, Medium, and Hard settings were categorized according to both the target distance and the required heading change, as shown in Table~\ref{tab:real_world_targets}. In all experiments, the robot was initialized facing the goal. A successful trial required the robot to score by kicking the ball into the goal with a single kick attempt. Score is defined as the ball completely crossing the goal line.

\textbf{Real World Dribbling Experiment.} For the real-world dribbling experiments showed in Table~\ref{tab:quantitative-results}, we design three experimental settings with increasing difficulty. The Easy difficulty level corresponds to the \textbf{Reaching Target Position} task. In this task, the robot was required to dribble the ball to a designated target location. The target was randomly generated for each trial at a distance of 4--5\,m from the robot and within an angular range of $\pm 30^\circ$ relative to its initial heading. The Medium difficulty level corresponds to the \textbf{Obstacle Traversal} task. In this experiment, the robot was required to dribble the ball to a designated target location while avoiding a predefined obstacle. The obstacle was positioned 3--4\,m away from the robot, and the target point was typically located approximately 3\,m beyond the obstacle. Depending on the obstacle configuration, successful traversal required the robot to execute a heading change of approximately $60^\circ$--$120^\circ$ during obstacle avoidance. The hard difficulty level corresponds to the \textbf{Slalom Dribbling} task. In this experiment, the robot was required to complete a continuous dribbling task by navigating through a sequence of target locations. Consecutive target points were separated by 2--3\,m, and transitioning between targets required a heading change of approximately $60^\circ$--$120^\circ$, consistent with the previous experimental setting. For each trial, the robot was required to visit 2--4 target points sequentially while maintaining control of the ball. For all experiments described above, the ball spawn distance was fixed at 0.6\,m.

For each of the three experimental settings, a trial was counted as successful
when the robot remained upright throughout execution and both the robot and
the ball finished within a 1\,m radius of the target point. The real-world experiments use the fixed deployment checkpoint. To ensure consistency, all trials were performed by the same operator using a joystick controller. This concrete end-state test supplements the task-level definition in the main-paper results table.

%% file: tables/appendix/dm_protocol.tex
\begin{table}[t]
\centering
\caption{Dribble Master's published dribbling protocol and our reproduction of it on DAVIS.}
\label{tab:dm-protocol}
\scriptsize
\begin{tabularx}{\linewidth}{>{\centering\arraybackslash}m{0.18\linewidth}>{\raggedright\arraybackslash}X>{\raggedright\arraybackslash}X}
\toprule
Field & Dribble Master~\cite{wang2025dribblemaster} & DAVIS reproduction \tabularnewline
\midrule
Robot / policy input & Booster T1; detector ball position + in-view flag & Noetix E1; depth image \tabularnewline
Command & global ball-velocity vector, issued by a human operator & direction only, scripted (potential field for reaching, fixed heading for turning) \tabularnewline
Reaching / obstacle & unreported distance; success = final error $<1$\,m & goal at $d\in\{3,5,8\}$\,m; success = minimum ball--goal distance $<\tau$, $\tau\in\{0.5,1.0,1.5\}$\,m; round ends at $0.5$\,m or after $\max(20,6d)$\,s \tabularnewline
Obstacle & physical; not perceived by the policy & virtual pole at the path midpoint; not perceived by the policy \tabularnewline
Turning & $\pm45^\circ/90^\circ$ on entering a $0.4$\,m circle at $(1.5,0)$\,m & $\pm30/45/60/90^\circ$ on $1.1$\,m of forward progress (where a straight path enters that circle) \tabularnewline
Trials & 5 rollouts per turn condition (sim); 15 per task (real) & 50 rounds per condition (sim), Wilson 95\% CI \tabularnewline
\bottomrule
\end{tabularx}
\end{table}

%% file: tables/appendix/dm_results.tex
\begin{table*}[t]
\centering
\scriptsize
\setlength{\tabcolsep}{4pt}
\caption{Target-reaching and single-obstacle results under the Dribble Master protocol (DAVIS simulation, $n{=}50$ rounds per row, Wilson 95\% CI). Success is reported at three tolerances, together with the time from first ball contact to the $1$\,m tolerance (mean $\pm$ SD over the rounds that got there, count in parentheses); a round ends when the ball first comes within $0.5$\,m. Dribble Master's reaching and obstacle values come from real-robot trials with $n{=}15$ at an unreported distance.}
\label{tab:dm-results}
\begin{tabular}{>{\centering\arraybackslash}m{0.22\linewidth}>{\centering\arraybackslash}m{0.08\linewidth}>{\centering\arraybackslash}m{0.12\linewidth}>{\centering\arraybackslash}m{0.12\linewidth}>{\centering\arraybackslash}m{0.12\linewidth}>{\centering\arraybackslash}m{0.18\linewidth}}
\toprule
Task & $d$ (m) & $\mathrm{RR}_{0.5}$ & $\mathrm{RR}_{1.0}$ & $\mathrm{RR}_{1.5}$ & time (s) \\
\midrule
\multirow[c]{3}{=}{\centering Target reaching} & 3 & 0.88 [0.76, 0.94] & 0.88 [0.76, 0.94] & 0.90 [0.79, 0.96] & $3.0\pm0.6$ (44) \\
 & 5 & 0.82 [0.69, 0.90] & 0.84 [0.71, 0.92] & 0.84 [0.71, 0.92] & $5.7\pm0.5$ (42) \\
 & 8 & 0.78 [0.65, 0.87] & 0.80 [0.67, 0.89] & 0.80 [0.67, 0.89] & $9.4\pm1.1$ (40) \\
\midrule
\multirow[c]{3}{=}{\centering Single obstacle} & 3 & 0.92 [0.81, 0.97] & 0.92 [0.81, 0.97] & 0.92 [0.81, 0.97] & $4.8\pm1.1$ (46) \\
 & 5 & 0.94 [0.84, 0.98] & 0.96 [0.87, 0.99] & 0.96 [0.87, 0.99] & $6.4\pm1.1$ (48) \\
 & 8 & 0.94 [0.84, 0.98] & 0.94 [0.84, 0.98] & 0.94 [0.84, 0.98] & $9.9\pm1.1$ (47) \\
\midrule
Dribble Master (real, $n{=}15$) & -- & -- & 0.87 [0.62, 0.96] (reach) / 0.93 [0.70, 0.99] (obstacle) & -- & 22.0 / 31.4 \\
\bottomrule
\end{tabular}
\end{table*}

\begin{table*}[t]
\centering
\scriptsize
\setlength{\tabcolsep}{4pt}
\caption{Turning results under the Dribble Master protocol (DAVIS simulation, $n{=}50$ rounds per condition). Achieved heading change is reported as mean $\pm$ SD over scored rounds, together with the bias of the mean. Dribble Master's turning values come from MuJoCo with 5 rollouts per condition.}
\label{tab:dm-turning-results}
\begin{tabular}{>{\centering\arraybackslash}m{0.18\linewidth}>{\centering\arraybackslash}m{0.10\linewidth}>{\centering\arraybackslash}m{0.10\linewidth}>{\centering\arraybackslash}m{0.12\linewidth}>{\centering\arraybackslash}m{0.12\linewidth}|>{\centering\arraybackslash}m{0.12\linewidth}>{\centering\arraybackslash}m{0.12\linewidth}}
\toprule
\multirow{2}{*}{Turn command} & \multicolumn{4}{c|}{DAVIS} & \multicolumn{2}{c}{Dribble Master} \\
\cmidrule(lr){2-5}\cmidrule(lr){6-7}
& scored & $\bar\theta\pm\sigma$ ($^\circ$) & bias (\%) & per-round (\%) & $\bar\theta$ ($^\circ$) & bias (\%) \\
\midrule
30$^\circ$ L & 40 & $+28.6\pm20.5$ & 4.8 & 52.9 & -- & -- \\
30$^\circ$ R & 39 & $-25.5\pm15.3$ & 15.1 & 41.5 & -- & -- \\
45$^\circ$ L & 44 & $+39.3\pm14.9$ & 12.6 & 27.6 & +43.58 & 3.16 \\
45$^\circ$ R & 44 & $-44.8\pm18.2$ & 0.5 & 32.1 & -46.44 & 3.20 \\
60$^\circ$ L & 43 & $+60.4\pm19.5$ & 0.6 & 26.3 & -- & -- \\
60$^\circ$ R & 41 & $-53.7\pm17.2$ & 10.6 & 23.3 & -- & -- \\
90$^\circ$ L & 42 & $+84.0\pm25.4$ & 6.7 & 18.8 & +88.06 & 2.16 \\
90$^\circ$ R & 43 & $-84.9\pm22.6$ & 5.6 & 19.7 & -87.98 & 2.24 \\
\midrule
Mean bias & & & 7.1 & & & 2.7 \\
\bottomrule
\end{tabular}
\end{table*}

%% file: appendix/appendix_C.tex
This section provides implementation details for the deployable observation interface, policy architecture, PPO / AMP-HIM optimization, and the auxiliary geometry and visibility losses used by DAVIS.

\subsection{Observation Space and Policy Architecture}
\label{app:observation-policy-architecture}

The actor input consists of a single aligned depth frame, a 5-step proprioceptive history, and an optional low-dimensional task command. The shooting policy uses an 81-D proprioceptive vector, while the dribbling policy uses an 84-D proprioceptive vector. The proprioceptive vector includes base angular velocity, projected gravity, relative joint positions, relative joint velocities, and the previous action. The depth image is encoded by a \texttt{LightDepthEncoder} into a 32-D latent feature. Both the actor and critic MLPs use hidden layers $[512,256,128]$ with ELU activations and empirical observation normalization.

During training, the critic additionally receives privileged quantities, including base linear velocity, foot contact states, terrain height scan, ball and goal geometry, auxiliary labels, and curriculum variables. These signals are used only for training-time value learning, reward computation, or auxiliary supervision, and are not actor inputs at deployment. The HIM estimator uses the proprioceptive history to produce a 3-D latent estimate of hidden robot state, which is concatenated with the actor features.

The depth latent is supervised by auxiliary geometry and visibility heads. The geometry heads predict local 3-D positions of the ball and goal, and the visibility head predicts whether each object is visible in the current depth frame. Ground-truth geometry and visibility are mixed with their predictions under the scheduled transition defined below. Table~\ref{tab:observation-network-config} summarizes the observation and network configuration, and Table~\ref{tab:ppo-amp-him-settings} lists the optimization settings.

\begin{table}[htbp]
    \centering
    \caption{Observation and network configuration.}
    \label{tab:observation-network-config}
    \begin{tabularx}{\linewidth}{>{\centering\arraybackslash}m{0.28\linewidth}>{\centering\arraybackslash}X}
        \hline
        Item & Configuration \\
        \hline
        Actor sensory input & Single-frame depth image + 5-step proprioceptive history \\
        Optional command & Low-dimensional task command if enabled \\
        Shooting proprioception & 81-D \\
        Dribbling proprioception & 84-D \\
        Depth encoder & \texttt{LightDepthEncoder} \\
        Depth latent & 32-D \\
        HIM estimate output & 3-D \\
        Actor MLP & $[512,256,128]$, ELU \\
        Critic MLP & $[512,256,128]$, ELU \\
        Actor normalization & Enabled \\
        Critic normalization & Enabled \\
        Auxiliary geometry heads & Goal 3-D, ball 3-D \\
        Auxiliary visibility head & 2-D: goal visible, ball visible \\
        \hline
    \end{tabularx}
\end{table}

\begin{table}[t]
    \centering
    \caption{PPO and AMP-HIM optimization settings.}
    \label{tab:ppo-amp-him-settings}
    \begin{tabularx}{\linewidth}{>{\centering\arraybackslash}m{0.28\linewidth}>{\centering\arraybackslash}X}
        \hline
        Item & Value \\
        \hline
        Rollout length & 24 steps per environment \\
        PPO epochs & 5 \\
        Mini-batches & 4 \\
        Learning rate & $1\times10^{-3}$, adaptive schedule \\
        Discount factor  & $\gamma=0.99$ \\
        GAE parameter &$\lambda=0.95$ \\
        PPO clip & 0.2 \\
        Entropy coefficient & 0.01 \\
        Value loss coefficient & 1.0 \\
        Max gradient norm & 1.0 \\
        Discriminator LR & $5\times10^{-6}$ \\
        Discriminator loss & Wasserstein loss \\
        Discriminator hidden layers & $[1024,512]$ \\
        AMP replay buffer & 200000 \\
        AMP reward coefficient & 0.8 \\
        AMP reward interpolation & 0.8 \\
        Auxiliary geometry loss & goal: 1.0, ball: 1.0 \\
        Visibility loss & goal: 0.2, ball: 0.2 \\
        \hline
    \end{tabularx}
\end{table}

The AMP discriminator uses a 136-D single-frame state composed of joint position, key-body positions, base linear velocity, base angular velocity, joint velocity, and foot-contact masks. 
We mask the head-related dimensions in the discriminator input because head yaw and pitch are optimized for active visual tracking rather than locomotion-style imitation.

\subsection{Auxiliary Geometry and Visibility Losses}
\label{app:auxiliary-head-losses}
Let $z_t^d$ denote the 32-D depth latent produced by the
\texttt{LightDepthEncoder}. For each task object $i$ (the ball, and
the goal in the shooting task), a geometry head predicts the object
geometry $\hat{\mathbf{y}}^i_t$, while a separate visibility head
predicts a visibility logit $\hat{v}^i_t$ from the shared latent,
\begin{equation}
\begin{gathered}
\hat{\mathbf{y}}^i_t
=
h^i_{\mathrm{geo}}(z_t^d),
\qquad
\hat{v}^i_t
=
h^i_{\mathrm{vis}}(z_t^d),
\\
p^i_t=\sigma(\hat{v}^i_t).
\end{gathered}
\end{equation}
The geometry is expressed in the head-yaw-aligned frame defined in
Sec.~III-C, with $x$ along the head-yaw forward direction, $y$ to the left,
and $z$ upward. Both auxiliary heads use only the current depth latent,
while head states are provided separately to the actor through
proprioception.
Simulator labels provide a target geometry $\mathbf{y}^{i*}_t$ and a
binary visibility label $v^{i*}_t$. Let $\mathcal{P}_i$ denote the
predefined structural points used for visibility evaluation. The
visibility label is defined as
\begin{equation}
v^{i*}_t
=
\mathbb{I}
\left[
\exists\,\mathbf{s}\in\mathcal{P}_i:
\operatorname{valid}
\left(
\mathbf{T}_{C\leftarrow R,t}\mathbf{s}
\right)
\right],
\end{equation}
where $\operatorname{valid}(\cdot)$ indicates that the transformed
point lies within the valid camera field of view and depth range.
For a single-point object, $\mathcal{P}_i$ contains only its reference
point; for a multi-point object, the object is considered visible if
any predefined structural point is valid. These structural points are
used only for visibility evaluation, while the geometry target remains
the object center.
Geometry is supervised only on visible frames, while visibility is
trained by binary cross-entropy on the logits:
\begin{equation}
\begin{gathered}
\mathcal{L}_{\mathrm{geo}}
=
\sum_i
v^{i*}_t\,
\ell_{\mathrm{geo}}
\left(
\hat{\mathbf{y}}^i_t,
\mathbf{y}^{i*}_t
\right),
\\
\mathcal{L}_{\mathrm{vis}}
=
\sum_i
\lambda_v\,
\mathrm{BCEWithLogits}
\left(
\hat{v}^i_t,
v^{i*}_t
\right),
\\
\mathcal{L}_{\mathrm{aux}}
=
\mathcal{L}_{\mathrm{geo}}
+
\mathcal{L}_{\mathrm{vis}}.
\end{gathered}
\end{equation}
The visibility BCE is the direct supervision for the visibility head;
the detached actor-side gate defined below prevents the policy gradient
from propagating through the visibility confidence.

\begin{figure}[htbp]
  \centering
  \includegraphics[width=\linewidth]{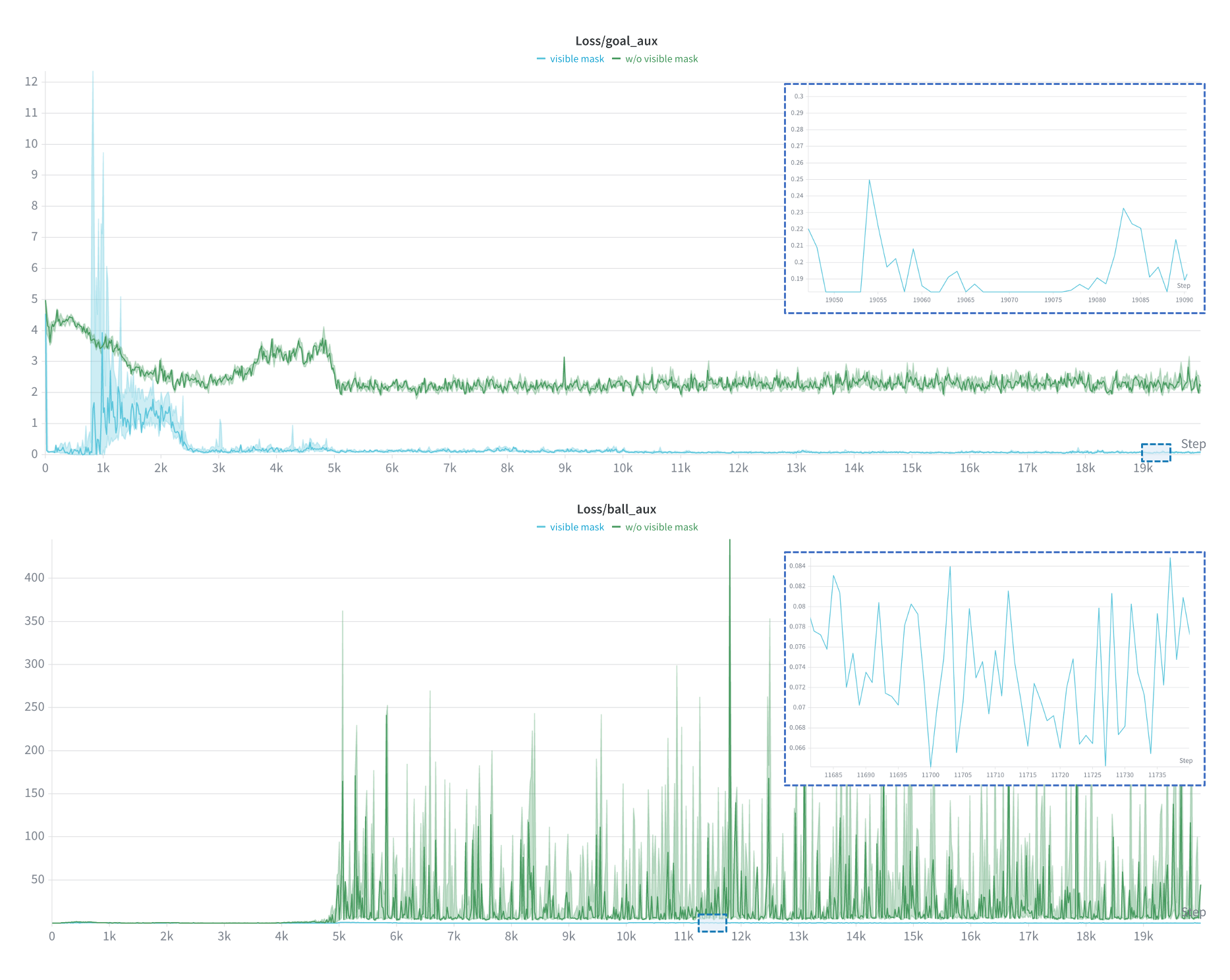}
  \caption{Loss curve of GT-annealing and w/o GT-annealing}
  \label{fig:loss_ablation}
  
\end{figure}

During training, the actor consumes geometry through a scheduled
ground-truth-to-prediction mixture, measured in policy updates $k$
(in code, world-size-normalized environment steps with
$k\!\approx\!\text{step}/24$). Let $\beta_k$ be the prediction weight
and $\alpha_k=1-\beta_k$ the ground-truth weight. Rather than a linear
decay, $\beta_k$ follows a monotone, concave $\sqrt{\cdot}$ ramp from
a warm-up update $k_0$ to an end update $k_1$,
\begin{equation}
\begin{gathered}
\beta_k
=
\max\!\Big(
\beta_{k-1},
\sqrt{
\mathrm{clip}
\big(
(k-k_0)/(k_1-k_0),\,0,\,1
\big)}
\Big),
\\
\alpha_k=1-\beta_k.
\end{gathered}
\end{equation}
That is, the actor uses pure ground-truth guidance for $k\le k_0$
and gradually transitions to pure prediction at $k_1$. The ramp is
ratcheted (non-decreasing) and is frozen whenever the auxiliary-loss
EMA exceeds a threshold before a grace update, so the actor is shifted
onto its own predictions only as they become reliable.
Geometry and visibility are mixed under the same training schedule:
\begin{equation}
\begin{gathered}
\mathbf{y}^{e,i}_t
=
(1-\beta_k)\mathbf{y}^{i*}_t
+
\beta_k\hat{\mathbf{y}}^i_t,
\\
p^{e,i}_t
=
(1-\beta_k)v^{i*}_t
+
\beta_k p^i_t.
\end{gathered}
\end{equation}
The mixed visibility confidence is converted into a soft, detached
actor-side gate,
\begin{equation}
c^i_t
=
\mathrm{clip}
\left(
\mathrm{stopgrad}
\left(
p^{e,i}_t
\right),
0.05,
1
\right),
\end{equation}
and the auxiliary actor feature is
\begin{equation}
\mathbf{a}^{\mathrm{aux}}_t
=
\left[
c^i_t\mathbf{y}^{e,i}_t,\;
c^i_t
\right]_i.
\end{equation}
The stop-gradient operation prevents the policy objective from
modifying the visibility head through the actor-side gating path.
At deployment, the GT-to-prediction mixture is disabled
($\beta_k=1$, equivalently $\alpha_k=0$), and all auxiliary actor
features are predicted from the depth branch.
Figure~\ref{fig:loss_ablation} shows the training curve of
GT-annealing.

%% file: appendix/appendix_D.1Goal-Directed_Shooting.tex
Shooting is trained as a goal-directed contact task. 
Task-specific training uses 1024 parallel environments, a 20~s episode length,
a 0.005~s physics step, and the same control decimation used by the shared
training and deployment configuration in Appendix~\ref{app:robot-system}.

Let $d_t$ denote the robot--ball distance, and define the ball-to-goal direction as
\begin{equation}
\mathbf{u}^{bg}_t=\frac{\mathbf{g}_t-\mathbf{b}_t}{\|\mathbf{g}_t-\mathbf{b}_t\|_2}.
\end{equation}
Near-ball alignment and contact terms are modulated by a smooth distance gate,
\begin{equation}
G(d_t;d_0,\tau)=\sigma\left(\frac{d_0-d_t}{\tau}\right),
\end{equation}
which gives larger weights to these terms as the robot approaches the ball. 
The shooting reward is organized into approach, active vision, ball--goal alignment, contact, terminal success, and safety regularization terms, as summarized in Table~\ref{tab:shooting-reward-summary}.

\begin{table*}[t]
  \centering
  \caption{Shooting reward summary.}
  \label{tab:shooting-reward-summary}
  \scriptsize
  \begin{tabularx}{\linewidth}{>{\centering\arraybackslash}m{0.25\linewidth}X>{\centering\arraybackslash}m{0.18\linewidth}}
\toprule
Reward Terms & Components & Weight \\
\midrule
\multirow{2}{*}{Approach and chase} 
    & Approach ball reward          & 3.0  \\
    & Adaptive velocity reward       & 2.0  \\
\midrule
\multirow{3}{*}{Active vision} 
    & Head track ball reward         & 3.0  \\
    & Keep in horizon reward         & -0.5 \\
    & Body follow head reward        & 2.0  \\
\midrule
\multirow{3}{*}{Ball--goal alignment} 
    & Robot behind ball reward       & 1.0  \\
    & Lateral kick line penalty      & 1.0  \\
    & Head yaw align reward          & 1.5  \\
\midrule
\multirow{3}{*}{Ball reward to goal} 
    & Ankle toward ball reward       & 2.0  \\
    & Ball velocity reward           & 4.0  \\
    & Ball velocity direction reward & 1.5  \\
\midrule
Sparse goal scored 
    & One-shot score reward          & 10.0 \\
\midrule
    Safety and style regularization &
    Unstable posture; Foot sliding; Excessive contact force; Joint limits; Joint acceleration; Action-rate changes; Upper body deviations; &
    task-dependent penalties \\
    \midrule
    AMP motion prior &
    AMP style reward. &
    coefficient 0.8 \\
    \bottomrule
  \end{tabularx}
\end{table*}

%% file: appendix/appendix_D.2_Directional_Dribbling.tex
For reproducibility, the checkpoint uses the 12-way command ring and controllable-band definitions in Sec.~III-D. The dribbling episode lasts 60~s with a 0.005~s physics step and the shared control decimation; training tightens the band half-width from 0.15~m to 0.10~m and the out-of-band dwell from 6.0~s to 3.5~s.

\begin{table*}[t]
  \centering
  \caption{Dribbling reward and regularization summary.}
  \label{tab:dribbling-reward-summary}
  \scriptsize
  \begin{tabularx}{\linewidth}{>{\centering\arraybackslash}m{0.25\linewidth}X>{\centering\arraybackslash}m{0.18\linewidth}}
    \toprule
    Group & Terms & Weight / setting \\
    \midrule
    \multirow{2}{*}{Commanded progress \& heading}
    & Ball command reward & 5.0 \\
    & Body command reward & 0.4 \\
    \midrule
    \multirow{2}{*}{Ball control}
    & Dense re-approach reward & 2.0 \\
    & Out-of-band penalty & -1.0 \\
    \midrule
    \multirow{2}{*}{Active vision}
    & Ball in view reward & 1.0 \\
    & Body follow head reward & 0.6 \\
\midrule
    Safety and style regularization &
    Unstable posture; Foot sliding; Excessive contact force; Joint limits; Joint acceleration; Action-rate changes; Upper body deviations; &
    task-dependent penalties \\
    \midrule
    Termination
    & Falls penalty; Bad-orientation penalty; Lose ball control penalty; & -1.0 \\
    \midrule
    AMP motion prior
    & Head-masked Wasserstein AMP style reward & coefficient 0.8, interpolation 0.8 \\
    \bottomrule
  \end{tabularx}
\end{table*}

%% file: appendix/appendix_E_sim2real.tex
This section records the remaining sim-to-real details beyond the deployment interface and policy settings already given in Appendices~\ref{app:robot-system} and~\ref{app:policy-architecture}: depth sensing and robot--ball--ground physical variation.

The simulated camera is mounted on the active head and renders \texttt{distance\_to\_image\_plane}. Both simulation and real deployment use a single $168\times80$ aligned depth frame, clipped to 0.3--5.0~m, normalized as $2D/5.0-1$, and zero-filled at invalid pixels. In simulation, this tensor is obtained by rendering a $168\times94$ pinhole depth image with calibrated ZED2i intrinsics and cropping it to the actor input size. The real controller applies the same cropping, range filtering, and normalization before actor inference. Temporal information is provided by proprioceptive history rather than by stacking depth frames.

During training, we randomize visual observations and physical interaction parameters. Depth perturbations are applied to the rendered tensor, while robot, ball, ground, and contact variations are applied through IsaacLab event terms at startup, reset, or fixed intervals. Table~\ref{tab:domain-randomization} summarizes the perturbation groups used for sim-to-real transfer.

\begin{table}[htbp]
\centering
\scriptsize
\setlength{\tabcolsep}{5pt}
\renewcommand{\arraystretch}{1.12}
\caption{Representative training randomization ranges for sim-to-real transfer.}
\label{tab:domain-randomization}
\begin{tabular}{@{}>{\centering\arraybackslash}m{0.48\linewidth}>{\centering\arraybackslash}m{0.38\linewidth}@{}}
\toprule
Perturbation & Range / value \\
\midrule
Valid depth range & 0.3--5.0~m \\
Distance-proportional depth noise & 0.01 \\
Gaussian depth-noise coefficient & 0.005 \\
Salt-and-pepper probability & 0.01 \\
Stripe dropout probability & 0.005 \\
Edge-drag probability & 0.01 \\
Rectangular mask probability & 0.1 \\
Rectangular mask size & 10--25\% \\
Ball-region occlusion probability & 0.2 \\
Ball-region occlusion radius & 2--10 px \\
Partial occlusion ratio & 45--85\% \\
Border near-field noise probability & 0.15 \\
Camera intrinsic perturbation & $\pm0.3$--$\pm0.5$ px \\
Camera translation perturbation & 0.02--0.04~m \\
Camera rotation perturbation & 0.025--0.060~rad \\
Joint-position offset & $\pm0.02$~rad \\
Base COM perturbation & $\pm0.05$~m \\
Base mass perturbation & $\pm5$~kg \\
Actuator stiffness / damping scale & 0.8--1.2 \\
Ball mass randomization & 0.36--0.56~kg \\
External push interval & 3--15~s \\
External linear velocity perturbation & $x,y\in[-1,1]$~m/s; $z\in[-0.2,0.2]$~m/s \\
External angular velocity perturbation & $[-0.78,0.78]$~rad/s \\
\bottomrule
\end{tabular}
\end{table}

%% file: appendix/additional_results.tex
This section contains additional empirical results that complement the main
paper. 

\subsection{Powerful Shot.}
We additionally study a close-range powerful-shot skill as another task-specific instantiation.
This skill uses the same depth-only actor interface and auxiliary ball-geometry prediction as the base skills; only the reset distribution, reward terms, and motion-prior schedule are changed.

We use the same near-ball reward gating as in the shooting task.
The reward follows a two-stage curriculum: the first stage rewards near-ball leg effort through a distance-weighted foot-speed term, and the second stage adds ball-outcome terms, including horizontal ball speed and displacement from the reset position.
The corresponding reward scales are ramped in sequentially during training.

\begin{figure}[htbp]
  \centering
  \includegraphics[width=\linewidth]{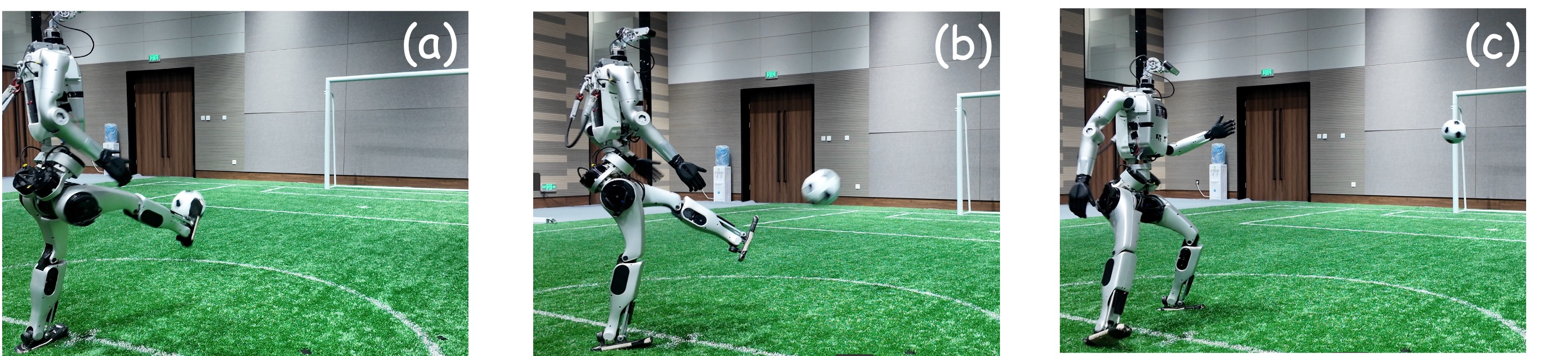}
  \caption{Power shoot real-world example}
  \label{fig:power shoot}
  
\end{figure}

An AMP kick-motion prior is activated once a sufficient fraction of parallel environments reaches the near-ball zone, regularizing the explosive leg swing while keeping the actor input unchanged.
This experiment tests whether the same depth-only framework can support high-speed striking from a close pre-kick stance without adding a separate runtime ball-localization or shooting controller. Figure~\ref{fig:power shoot} shows the real-world result.

\subsection{Dribbling while Avoiding Obstacles.}
This skill highlights the extensibility of our depth-predicted auxiliary heads:
extending directional dribbling to avoid obstacles requires no architectural change.
Following the scalable visibility-aware auxiliary geometry of
Sec.~\ref{sec:visibility-aux}, the obstacle simply enters the task object set as one
more object, adding a single indexed geometry\,/\,visibility head that predicts its
relative position and visibility from the shared depth latent---with no privileged
obstacle input and no change to the actor interface or control loop.

\begin{figure}[htbp]
  \centering
  \includegraphics[width=\linewidth]{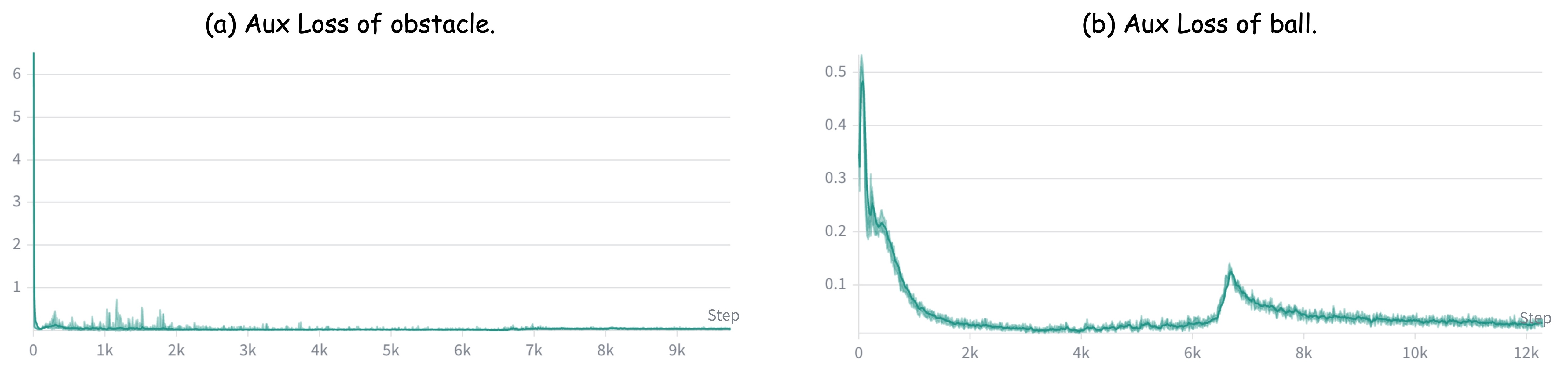}
  \caption{Auxiliary loss for the ball and obstacles. The prediction heads for both objects converge during training.}
  \label{fig:loss}
  
\end{figure}

Avoidance is induced through the reward rather than a separate controller: an
artificial potential field deflects the commanded direction $\mathbf{d}^{\mathrm{cmd}}_t$
by a tangential repulsion from the nearest in-path obstacle,
\begin{equation}\label{eq:apf}
\tilde{\mathbf{d}}^{\mathrm{cmd}}_t=
\frac{\mathbf{d}^{\mathrm{cmd}}_t+\kappa\,w_t\,a_t\,\mathbf{t}_t}
{\big\lVert\,\mathbf{d}^{\mathrm{cmd}}_t+\kappa\,w_t\,a_t\,\mathbf{t}_t\,\big\rVert},
\end{equation}
where $\mathbf{t}_t$ is the unit normal to the command on the obstacle-clearing side
($\mathbf{t}_t\!\cdot\!\mathbf{u}_t\le0$, with $\mathbf{u}_t$ the ball-to-obstacle
bearing), $a_t=\max(\mathbf{u}_t\!\cdot\!\mathbf{d}^{\mathrm{cmd}}_t,\,0)$ activates the
repulsion only for obstacles ahead, $w_t\in[0,1]$ is a proximity weight that vanishes
beyond an influence radius, and $\kappa$ is the steering gain. Every direction-aware
dribble reward tracks $\tilde{\mathbf{d}}^{\mathrm{cmd}}_t$ in place of
$\mathbf{d}^{\mathrm{cmd}}_t$, so obstacle avoidance emerges from the same
velocity-direction tracking objective with no change to the reward structure. We show detailed loss curve of the aux loss in Figure \ref{fig:loss}.

\subsection{Recovery upon Losing the Ball.}

We also test whether the chase policy can recover after the ball temporarily
leaves the head-mounted field of view. The recovery behavior is not implemented
as a separate search controller. Instead, the same active-vision policy is shaped
by a visibility flag and an invisible-ball timer: when the ball is visible, the
robot continues pursuit; when it is lost, the policy is penalized for prolonged
invisibility and for moving too fast without reliable visual evidence. In the
submitted configuration, the search penalty starts after \(0.5\,\mathrm{s}\) of
continuous invisibility with weight \(0.2\), and the invisible-speed penalty
limits body-frame forward speed around \(0.5\,\mathrm{m/s}\) with weight \(0.6\).

Recovery is driven by the active head and body-following rewards. The head
reward tracks the desired yaw/pitch angles toward the ball in the robot body
frame, while the body-follow-head reward encourages the torso to realign with
the recovered viewing direction. These rewards are introduced gradually: the
head reward starts at 12k steps and ramps over 36k steps, and the body-following
reward starts at 48k steps and ramps over 48k steps. This produces a closed-loop
search-and-reacquire behavior while preserving the same depth-only actor
interface used by the main soccer skills.
